\documentclass{article}

  \usepackage[preprint]{neurips_2026}

\usepackage{amsmath}

\usepackage[utf8]{inputenc} 
\usepackage[T1]{fontenc}    
\usepackage{hyperref}       
\usepackage{cleveref}       
\usepackage{url}            
\usepackage{booktabs}       
\usepackage{amsfonts}       
\usepackage{nicefrac}       
\usepackage{microtype}      
\usepackage{xcolor}         

\usepackage{relsize}
\usepackage{caption}
\usepackage{subfigure}
\usepackage{wrapfig}
\usepackage{pgfplots}
\pgfplotsset{compat=1.18}
\usepackage{pgfplotstable}
\usetikzlibrary{positioning, calc, matrix}
\usepackage{tikz}
\usetikzlibrary{shapes, arrows.meta, positioning, calc, fit, backgrounds}

\usepackage{rotating}
\usepgfplotslibrary{groupplots}

\newsavebox{\leftimagebox}

\def\x{\mathbf{x}}

\def\our{SEC-conf}
\def\fresh{FreSh}
\def\siren{{\em Siren}}
\def\ffeatures{{\em Fourier}}
\def\finer{{\em Finer}}
\def\wire{{\em Wire}}

\def\chest{Chest X-Ray}
\def\ffhqcropped{FFHQ-1024}
\def\ffhq{FFHQ-wild}
\def\kodak{Kodak}
\def\art{Wiki Art}

\newcommand{\real}[0]{\mathbb{R}}

\newcommand{\inherentfreq}{SEC} 
\newcommand{\Inherentfreq}{Energy centroid}
\def\inherentFreqMath{\text{SEC}}
\newcommand{\freqparam}{embedding frequency} 
\def\power{\mathcal{E}} 
\def\normalPower{\Tilde{\power}} 
\def\freqParamMath{F}

\DeclareMathOperator*{\argmin}{arg\,min}

\title{Spectral Energy Centroid: a Metric for Improving Performance and Analyzing Spectral Bias in Implicit Neural Representations}

\author{
  Tomasz Dądela\thanks{Equal contribution.} \\
  Jagiellonian University \\
  Kraków, Poland \\
  \And
  Adam Kania\footnotemark[1] \\
  GSK.ai \\
  Baar, Switzerland \\
  \And
  Maciej Rut\footnotemark[1] \\
  Jagiellonian University \\
  Kraków, Poland \\
  \And
  Przemysław Spurek \\
  Jagiellonian University \\
  IDEAS \\
  Kraków, Poland \\
}

\begin{document}

\maketitle

\begin{abstract}
    Implicit Neural Representations (INRs) model continuous signals using
multilayer perceptrons (MLPs),
enabling compact, differentiable, and high-fidelity representations of
data across diverse domains.
However, due to the low-frequency bias of MLPs that prevents effective
learning of small details,
the model's frequency must be carefully tuned through the embedding layer.
Prior work established that this tuning can be performed before
training based on the target signal,
but it did not account for the significant effect of model depth, indicating that our understanding of the relationship between frequency and INR performance remains limited.
To gain insights into this relationship,
we utilize the Spectral Energy Centroid (SEC) metric
that quantifies the frequency of target images and
the spectral bias of INR models.
We show that SEC is a versatile tool
for INR analysis, demonstrating its utility across
three tasks: (1)~a data-driven strategy (SEC-Conf)
for hyperparameter selection that outperforms
existing heuristics and is robust to model depth,
(2)~a reliable proxy for signal complexity, and
(3)~effective alignment of spectral biases across
diverse INR architectures.
\end{abstract}

\begin{wrapfigure}{r}{0.47\textwidth}
    \centering
   \vspace{-13pt}
    \includegraphics[width=0.44\textwidth]{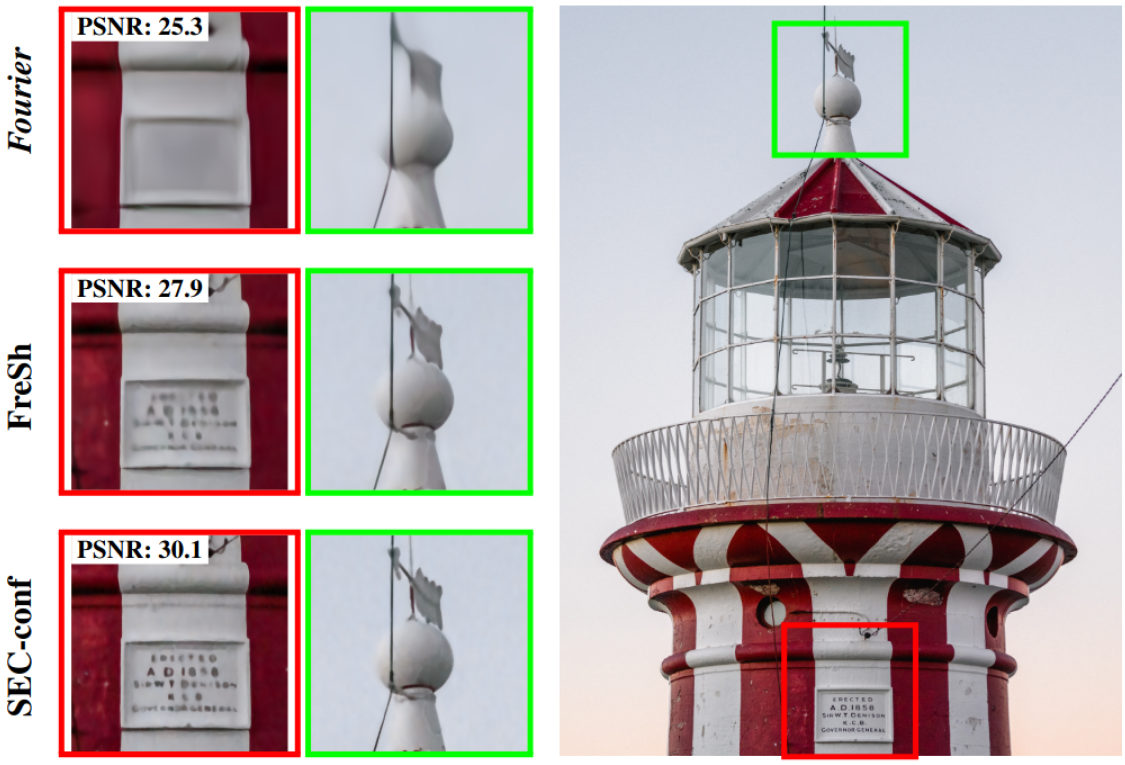}

    \caption{Visual comparison of performance of a large (4-layer) \ffeatures{} model after 1000 steps.
    The left block shows crops of the full image (right) for the default configuration of \freqparam{},
        and ones selected by \fresh{} and \our{} (\our{} is an example application of \inherentfreq{}).
    Leveraging \Inherentfreq{}s (\inherentfreq{}), we achieve improved performance, with \our{} resulting in the sharpest image.
    }
    \label{fig:motivation}
    \vspace{-29pt}
\end{wrapfigure}

Implicit Neural Representations (INRs) have become a vital tool for scene
    modeling, robotics, and generative tasks \cite{xie2022neural}, representing signals as
    continuous functions parameterized by MLPs.
    However, standard MLPs suffer from a low-frequency bias, making it difficult
    to capture fine details \cite{DeepLearningFourierAnalysis, SpectralBiasOfNN}.
    To address this, architectures like \siren{} \cite{SIREN}, \ffeatures{}
    \cite{FourierFeatures}, and \finer{} \cite{liu2024finer}
    introduce architectural inductive biases that facilitate the learning of
    fine details, in particular through a
    high-frequency input mapping controlled by an \freqparam{} parameter.

    However, in practice, selecting optimal \freqparam{} parameter values is
    computationally expensive.
    Optimal settings vary significantly across target images \cite{fresh},
    forcing practitioners to resort to costly hyperparameter sweeps or to rely
    on default values from original publications.
    This challenge is compounded in architectures like \finer{} or \wire{}
    \cite{WIRE}, where multiple frequency-related hyperparameters jointly
    influence performance, making it difficult to know which ones to tune.

    The \fresh{} \cite{fresh} method represents the first attempt to mitigate
    this issue and simplify \freqparam{} parameter selection for INR models by
    aligning the frequency of the model with the target image.
    \fresh{} assumes that for optimal learning, frequencies present in the
    image and the untrained model should be the same and aligns them by
    selecting an \freqparam{} parameter that minimizes Wasserstein distance.
    This approach, while effective in certain settings, does not generalize to
    some particularly high-frequency models such as \wire{} and degrades in
    performance when model depth is modified.
    This degradation suggests a flaw in the assumption made by \fresh{}
    regarding the alignment of model and target signal frequencies.
    Based on our experiments, the relationship between the frequencies of the
    target signal and the model is not direct, but can be learned from data.

    In this work, we investigate the spectral bias of INRs
    and its impact on performance across different target
    signals.
    To quantify the frequency content of target images and
    the spectral bias of INR models, we utilize the Spectral
    Energy Centroid metric (\inherentfreq{}).
    \textbf{
    We show that \inherentfreq{} is a general-purpose tool
    for INR spectral analysis, demonstrating its utility
    across three tasks.}
    We demonstrate that model architecture influences
    spectral bias, a dependency that undermines existing
    heuristics like \fresh{} which doesn't have a mechanism to account for model size.
    To address this, we propose a robust, data-driven strategy for embedding
    layer configuration (\our{}) that effectively handles varying model sizes,
    as shown in \cref{fig:motivation}.
    Furthermore, we provide a framework for aligning spectral biases across
    diverse architectures, demonstrating that such alignment can enable older
    models to match the performance of newer ones in specific settings.
    Finally, we validate \inherentfreq{} as a reliable proxy for signal
    complexity.
    Our main contributions are as follows:

    \begin{itemize}
        \item We utilize \inherentfreq{} to quantify the frequency of images
        and the spectral bias of INR models, revealing a strong dependency
        between model depth and spectral bias.
        \item We propose a data-driven strategy for hyperparameter selection
        (called \our{}) and alignment of spectral bias across architectures,
        outperforming existing heuristics.
        \item We demonstrate that \inherentfreq{} serves as a reliable proxy for signal complexity, strongly correlating with reconstruction quality.
    \end{itemize}

    \section{Related works}

    \textbf{Spectral bias} refers to the tendency of neural networks to learn
    low-frequency components of a signal more rapidly than high-frequency ones
    \citep{DeepLearningFourierAnalysis, SpectralBiasOfNN, ronen2019convergence,
    WalshHadamardRegularizer}.
    This phenomenon has been observed not only in MLPs but also in more complex
    architectures, such as diffusion models \cite{wang2025analytical}.
    Spectral bias is often cited as one explanation for the generalization
    ability of neural networks \cite{ronen2019convergence}.
    However, it poses a significant challenge for INRs,
    as it suppresses high-frequency signals.

    \textbf{Implicit Neural Representations} (INRs) represent signals - ranging
    from images and videos \cite{mihajlovic2023resfields} to 3D scenes
    \cite{Nerf} and tabular data \cite{tabINR} - as continuous functions
    parameterized by MLPs. In this study,
    we focus on INRs for image representation, as their moderate
    computational requirements enable us to conduct a large-scale benchmark
    across diverse architectures and configurations.
    INRs learn a mapping from spatial or temporal coordinates to signal values
    (e.g., color or density).
    Unlike in standard learning scenarios, INR training involves dense
    supervision (e.g., every pixel in an image) with the explicit goal of
    high-fidelity signal reconstruction.
    In this overfitting regime, spectral bias becomes a critical bottleneck,
    hindering the capture of fine details and removing it to
    enable learning of a broad
    range of frequencies is a central focus in INR research \cite{SIREN,
    FourierFeatures, WIRE, liu2024finer, fresh}. 
    This can be partially achieved through deeper models, 
    as it was theoretically shown depth expands the set of representable frequencies,
    though with diminishing magnitude for higher
    harmonics \cite{inr_dictionaries2022}.
    However, many state-of-the-art
    approaches primarily solve this issue with a high-frequency
    embedding in the input layer, governed by a hyperparameter we refer to as the
    {\it \freqparam{} parameter}. 

    \textbf{\siren{}} \cite{SIREN} is one of the earliest and most influential
    INR architectures.
    It employs sinusoidal activations throughout the network, enabling it to
    represent high-frequency signals effectively.
    Its first layer applies a scaled sinusoidal mapping, $\sin(\omega_s W
    \mathbf{x} + \mathbf{b})$, where $\omega_s$ serves as the primary frequency
    parameter controlling the network's bandwidth\footnote{We deviate from the
    original notation to avoid collisions between \siren{}, \finer{}, and
    \wire{}.}.
    Although all layers use sine activations, mainly the first layer affects
    the frequency bias, as later layers use relatively small scales.

    \textbf{\ffeatures{} features} \cite{FourierFeatures} utilize a fixed
    high-frequency embedding of the input coordinates, typically followed by an
    MLP with ReLU activations.
    The embedding utilizes frequencies
    sampled from a Gaussian distribution, where the standard deviation $\sigma$
    directly controls the spectral bias (see \cref{eq:fourier_features}).

    \textbf{\wire{}} \cite{WIRE} employs a continuous complex Gabor wavelet as
    its activation, which is governed by two hyperparameters: the frequency parameter
    $\omega_w$ and a spatial compactness parameter $s_0$.
    Unlike in \siren{} or \ffeatures{}, this activation is re-used at every
    layer without any change to its parameters, leading to a significant
    frequency increase throughout the model (see
    \cref{fig:freq_vs_width_and_layers} and \cref{eq:wire_activation}).
    In this work, we focus on $\omega_w$ as the relevant \freqparam{}
    parameter, while fixing $s_0=10$ across all experiments.

    \textbf{\finer{}} \cite{liu2024finer} employs a variable-periodic activation
    function whose frequency increases with the distance from the origin.
    These activations are governed by two parameters that jointly influence the
    frequency behavior, making the model more challenging to tune in practice.
    Specifically, \finer{} introduces a \siren{}-like frequency parameter
    $\omega_f$ together with an additive bias term (see \cref{eq:finer_layer}).

    \textbf{Initialization and configuration strategies} address spectral bias
    through specialized weight initialization, meta-learning, or automated
    hyperparameter selection.
    Meta-learning methods \cite{rajeswaran2019meta, tancik2021learned} learn
    initializations that facilitate rapid convergence, though they entail
    significant computational overhead.
    Recent efficient initialization schemes, such as \fresh{}, TUNER \cite{tuner}, and
    WINNER \cite{winner}, provide alternatives that avoid this pre-training cost.
    However, TUNER and WINNER are restricted in their applicability,
    as they are compatible only with \siren{}.
    \fresh{} \cite{fresh} automates the selection of frequency
    parameters for INRs.
    Operating on the premise that the spectral bias of an untrained network is
    characterized by the frequency spectrum of its output at initialization,
    \fresh{} selects hyperparameters by minimizing the Wasserstein distance
    between the spectrum of the initialized model's output and that of the
    target signal.
    Although effective in many scenarios, we demonstrate that this heuristic
    overlooks the influence of model depth and width on spectral bias, often
    resulting in suboptimal configurations for non-standard model sizes.
    Additionally, the full frequency spectrum is ill-suited for further analysis
    due to its multidimensional nature, making its applications limited.

    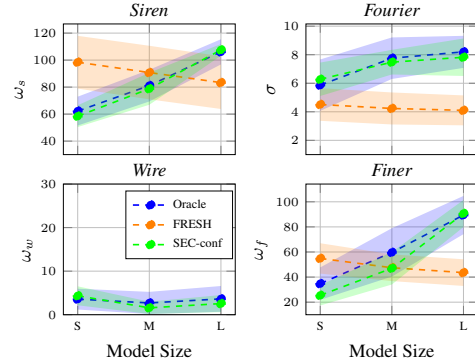
\begin{wrapfigure}{r}{0.45\textwidth}
        \centering
       \vspace{-50pt}
        \resizebox{0.46\textwidth}{!}{
            \begin{tikzpicture}
    \begin{groupplot}[
        group style={
            group size=2 by 2,
            horizontal sep=1.2cm,
            vertical sep=0.5cm,
            xlabels at=edge bottom,
            xticklabels at=edge bottom,
        },
        width=4.5cm,
        height=3.8cm,
        xtick={0,1,2},
        xticklabels={S, M, L},
        grid=major,
        xlabel={Model Size},
        ylabel style={font=\small, yshift=-1mm},
        xlabel style={font=\small},
        tick label style={font=\tiny},
        title style={font=\small, yshift=-2mm},
        tick style={font=\tiny},
        cycle list name=color list,
    ]
    \nextgroupplot[title={\siren{}}, ylabel={$\omega_s$}, ymin=30]
        \addplot [fill=blue, opacity=0.2, draw=none, forget plot] coordinates {(0,72.90432715490094) (1,92.86174527467618) (2,115.43601367040495) (2,96.78620855181728) (1,69.36047694754605) (0,51.5401172895435)} -- cycle;
        \addplot [dashed,blue, mark=*, thick] coordinates {(0,62.22222222222222) (1,81.11111111111111) (2,106.11111111111111)};
        \addplot [fill=orange, opacity=0.2, draw=none, forget plot] coordinates {(0,117.98472902778668) (1,110.62142379208355) (2,102.87872849204302) (2,63.78793817462363) (1,70.934131763472) (0,78.9041598611022)} -- cycle;
        \addplot [dashed,orange, mark=*, thick] coordinates {(0,98.44444444444444) (1,90.77777777777777) (2,83.33333333333333)};
        \addplot [fill=green, opacity=0.2, draw=none, forget plot] coordinates {(0,66.50476633713166) (1,90.65995003196397) (2,111.87720503026682) (2,103.23390608084429) (1,66.89560552359157) (0,50.38412255175723)} -- cycle;
        \addplot [dashed,green, mark=*, thick] coordinates {(0,58.44444444444444) (1,78.77777777777777) (2,107.55555555555556)};
    \nextgroupplot[title={\ffeatures{}}, ylabel={$\sigma$}, ymin=1]
        \addplot [fill=blue, opacity=0.2, draw=none, forget plot] coordinates {(0,7.666791506282588) (1,9.211354584091309) (2,9.323797734610016) (2,7.076202265389982) (1,6.299756527019802) (0,4.066541827050744)} -- cycle;
        \addplot [dashed,blue, mark=*, thick] coordinates {(0,5.866666666666666) (1,7.7555555555555555) (2,8.2)};
        \addplot [fill=orange, opacity=0.2, draw=none, forget plot] coordinates {(0,5.6551670188795535) (1,5.355630335823205) (2,5.149879607865223) (2,3.050120392134776) (1,3.111036330843461) (0,3.3670552033426686)} -- cycle;
        \addplot [dashed,orange, mark=*, thick] coordinates {(0,4.511111111111111) (1,4.233333333333333) (2,4.1)};
        \addplot [fill=green, opacity=0.2, draw=none, forget plot] coordinates {(0,7.448850250590318) (1,8.34189098532392) (2,9.134033542049302) (2,6.510410902395142) (1,6.613664570231637) (0,5.106705304965238)} -- cycle;
        \addplot [dashed,green, mark=*, thick] coordinates {(0,6.277777777777778) (1,7.477777777777778) (2,7.822222222222222)};
    \nextgroupplot[title={\wire{}}, ylabel={$\omega_w$}, ymin=0, ymax=30]
        \addplot [fill=blue, opacity=0.2, draw=none, forget plot] coordinates {(0,5.9714857179558205) (1,5.245981632280586) (2,6.591658094622808) (2,0.6750085720438586) (1,0.15401836771941513) (0,1.1840698375997354)} -- cycle;
        \addplot [dashed,blue, mark=*, thick] coordinates {(0,3.577777777777778) (1,2.7) (2,3.6333333333333333)};
        \addplot [fill=green, opacity=0.2, draw=none, forget plot] coordinates {(0,6.522790952260603) (1,2.9918147731282265) (2,4.516472052532124) (2,0.5946390585789869) (1,0.16374078242732892) (0,2.166097936628285)} -- cycle;
        \addplot [dashed,green, mark=*, thick] coordinates {(0,4.344444444444444) (1,1.5777777777777777) (2,2.5555555555555554)};
    \nextgroupplot[title={\finer{}}, ylabel={$\omega_f$}, legend style={at={(-1.05,0.95)}, anchor=north west, font=\tiny}, legend cell align=left,ymin=10]
        \addplot [fill=blue, opacity=0.2, draw=none, forget plot] coordinates {(0,47.693039873695234) (1,78.9487244616477) (2,104.68492212069363) (2,74.20396676819526) (1,40.38460887168563) (0,21.862515681860323)} -- cycle;
        \addplot [dashed,blue, mark=*, thick] coordinates {(0,34.77777777777778) (1,59.666666666666664) (2,89.44444444444444)};
        \addlegendentry{Oracle}
        \addplot [fill=orange, opacity=0.2, draw=none, forget plot] coordinates {(0,66.99742839985548) (1,58.10373496525263) (2,54.188460927756616) (2,32.9226501833545) (1,36.78515392363626) (0,42.780349377922285)} -- cycle;
        \addplot [dashed,orange, mark=*, thick] coordinates {(0,54.888888888888886) (1,47.44444444444444) (2,43.55555555555556)};
        \addlegendentry{FRESH}
        \addplot [fill=green, opacity=0.2, draw=none, forget plot] coordinates {(0,33.539540908758) (1,59.93458042560782) (2,101.4557581197049) (2,79.87757521362845) (1,34.06541957439218) (0,17.349347980130887)} -- cycle;
        \addplot [dashed,green, mark=*, thick] coordinates {(0,25.444444444444443) (1,47.0) (2,90.66666666666667)};
        \addlegendentry{SEC-conf}
    \end{groupplot}
\end{tikzpicture}
        }
        \caption{Mean \freqparam{} parameter values selected using \fresh{} and \our{}
        on a benchmark of images sampled from the LIU4K-v2 dataset. Y-axis is scaled to
        reflect the range of tested parameter values.
        Optimal \freqparam{} parameter values generally increase with model size (oracle).
        \wire{} is an exception, consistently requiring lower \freqparam{} values to compensate
        for its naturally high spectral bias.
        While \fresh{} tends to select smaller values when network size increases,
            our method correctly tracks the oracle.}
        \label{fig:sigma_vs_model_size}
       \vspace{-10pt}
    \end{wrapfigure}

      \section{Motivation}
    \label{sec:motivation}

    Implicit Neural Representations require careful tuning of the \freqparam{}
    parameter to capture high-frequency components of a signal.
    While previous work has established that the optimal embedding
    configuration is highly dependent on the target image \cite{fresh}, the
    influence of the network architecture itself remains under-explored, as we
    show in this section.

    We conduct a controlled experiment using models of varying sizes:
    small (S, $2\times128$), medium (M, $3\times256$), and large (L,
    $4\times512$).
    For each configuration, we perform a dense grid search over frequency
    parameters, training on a diverse set of 90 images.
    The grid search is described in more detail in \cref{sec:experiments}.
    We report the best performing configurations across several architectures in \cref{fig:sigma_vs_model_size},
    along with those selected by \fresh{} and \our{}.
    We observe that the optimal value of \freqparam{} systematically increases
    with model size, indicating that the network's capacity to utilize
    high-frequency input signals increases with its size.
    This dependency poses a challenge for \fresh{} \cite{fresh}, which fails to
    adapt to this architectural shift and doesn't follow the oracle.
    Specifically, while the optimal parameter values increase with model size,
    the embedding frequencies selected by \fresh{} decrease or remain
    relatively constant.
    This results in frequency parameters that are too high for smaller models
    and too low for larger ones.
    We attribute this failure to the implicit assumption that the spectral bias
    of an untrained INR remains static across different depths and widths.
    To address this, we propose using \inherentfreq{}s to capture the frequency
    characteristics of images and models as a scalar value,
    which allows us to better understand spectral bias of various architectures
    and frequencies of target signals. Utilizing SEC, we create
    robust initialization strategies for \freqparam{} parameters, 
    align spectral bias between architectures and analyze image complexity.

    \section{Spectral energy centroid}
    \label{sec:method}

        \begin{figure}
        \centering
        \resizebox{0.95\textwidth}{!}{
            \includegraphics{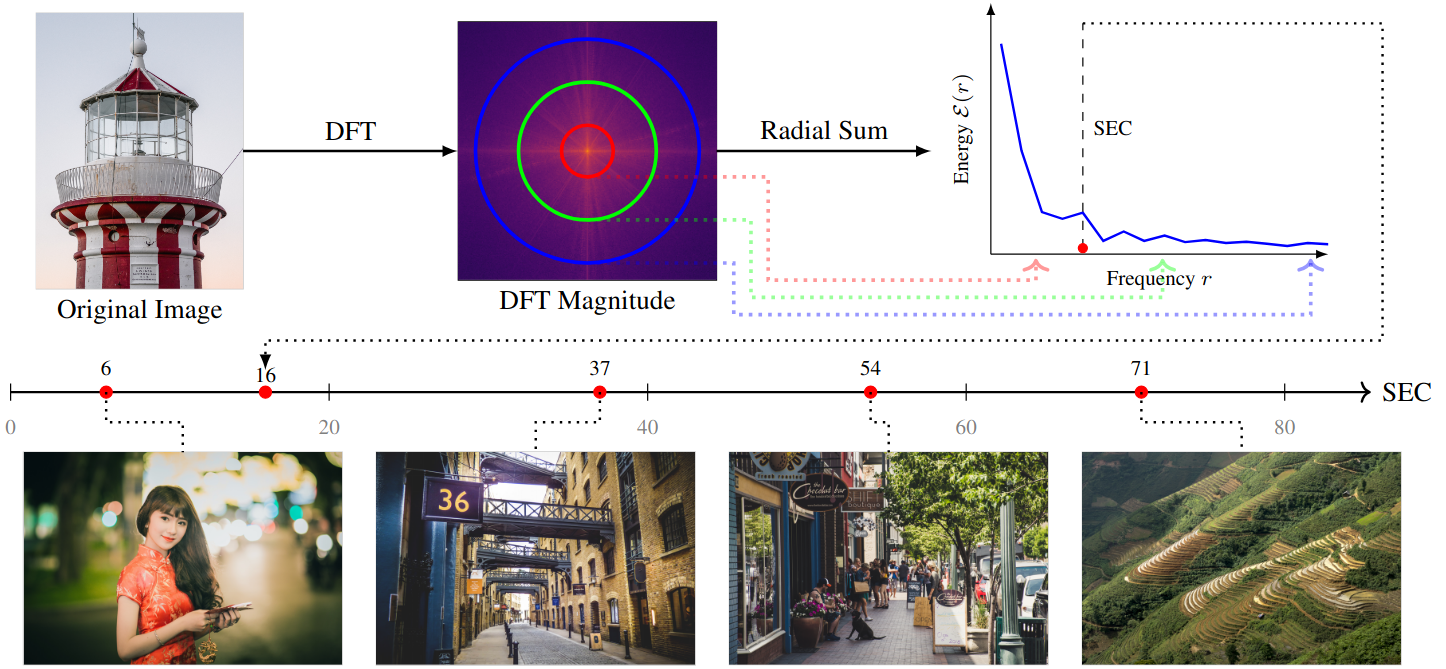}
        }
        \vspace{3pt}
        \caption{Visualization of the \inherentfreq{} calculation pipeline (top) and sample images with varying \inherentfreq{} values (bottom).
        To compute SEC, images are transformed via DFT and
        radial summation to obtain an energy spectrum, from which the centroid
        is computed.
        Our metric is correlated with visual complexity, as \inherentfreq{} increases
        images transition from smooth, blurry
        backgrounds to complex structures with high-frequency textures, such as
        dense foliage.}
        \label{fig:inherent_freq_pipeline}
    \end{figure}

    Motivated by the observation that architectural changes can substantially
    alter the spectral bias of an INR,
    our goal is to obtain a representation of frequency that can be easily
    interpreted and used to improve our understanding of INR models.
    We first establish the foundations of spectral analysis and then
    introduce a metric that achieves this goal, Spectral Energy Centroid (\inherentfreq{}),
    and its applications.

    \subsection{Preliminaries}
    \textbf{Discrete Fourier Transform} (DFT) is used to analyze the frequency
    content of images.
    Let $A \in \real^{C \times N \times M}$ denote an image with $C$ channels.
    The DFT of a channel $c$ is:
    \begin{equation}
        \mathcal{F}_{u,v}(A_c) = \frac{1}{NM} \sum_{n=0}^{N-1} \sum_{m=0}^{M-1}
        A_{c,n,m} e^{-i2\pi (\frac{un}{N} + \frac{vm}{M})},
    \end{equation}
    where $u \in \{0, \dots, N-1\}$ and $v \in \{0, \dots, M-1\}$ are frequency
    indices.
    To facilitate further analysis and follow standard practice, we shift the
    frequency indices to be centered at zero.
    We define a shift operator that maps indices from a grid centered at $(0,
    0)$ to the standard DFT grid:
    \begin{equation}
        \operatorname{Shift}(u,v)
        =
        \Bigl(
        u \bmod N,
        \;
        v \bmod M
        \Bigr).
    \end{equation}
    This allows us to consider frequencies on a grid $I = \{(u, v) \in
    \mathbb{Z}^2 : -\frac{N}{2} \le u < \frac{N}{2}, -\frac{M}{2} \le v <
    \frac{M}{2} \}$, where each point around $(0,0)$ corresponds to
    low-frequency components of the DFT.
    Following \fresh{}, we consider $\operatorname{Shift}(0,0)=0$, as the
    constant component of a signal can be easily modeled by an INR, yet it can
    be a major DFT component.

    \textbf{Energy spectrum.}
    We define the \textbf{energy spectrum}, $\power$, as a vector representing how the
    energy of the original image is distributed across radial frequencies.
    Specifically, for an image $A$, the $r$-th element of $\power$ captures the
    energy contained in frequencies with magnitude $r$:
    \begin{equation}
        \label{eq:spectrum_full}
        \power(A, r) = \sum_{c=0}^{C-1} \sum_{\substack{(i,j) \in I \\
        \lfloor\sqrt{i^2+j^2}\rfloor=r}} |\mathcal{F}_{\operatorname{Shift}(i,
        j)}(A_c)|^2.
    \end{equation}
    This characterization leverages signal energy, defined as the squared
    magnitude of the Fourier coefficients.
    By Parseval’s theorem \cite{oppenheim1997signals}, the total energy is
    preserved between the spatial and frequency domains, which
    ensures that our spectral analysis faithfully reflects the information
    content of the original signal.

    \begin{wrapfigure}{r}{0.50\textwidth}
    \vspace{-42pt}
    \centering
    \resizebox{\linewidth}{!}{ 
\begin{tikzpicture}[
    font=\sffamily\small,
    >={Latex[length=2mm, width=2mm]},
    node distance=0.7cm and 0.8cm,
    block/.style={rectangle, draw=gray!80, thick, rounded corners=3pt, fill=white, inner sep=5pt, align=center},
    arrow/.style={->, thick, gray!50!black},
]

\tikzset{
    pics/spectrum/.style={
        code={
            \draw[->, thin] (0,0) -- (1.2,0) node[right, scale=0.5] {$r$};
            \draw[->, thin] (0,0) -- (0,0.8) node[above, scale=0.5] {$E$};
            \draw[thick, blue!70!black, smooth, domain=0.01:1.1, samples=40] 
                plot (\x, {0.8*exp(-3.2*\x)});
            \draw[red, dashed, thick] (0.32,0) -- (0.32,0.4);
            \node[red, scale=0.55, anchor=north] at (0.32,0) {SEC};
        }
    },
    pics/neuralnet/.style={
        code={
            \foreach \y [count=\i] in {0.1, 0.5, 0.9} 
                \node[circle, fill=blue!30, draw=blue, inner sep=2pt] (in\i) at (0,\y) {};
                
            \foreach \y [count=\i] in {0.0, 0.33, 0.66, 1.0} 
                \node[circle, fill=green!30, draw=green!60!black, inner sep=2pt] (hid\i) at (0.9,\y) {};
                
            \foreach \y [count=\i] in {0.1, 0.5, 0.9} 
                \node[circle, fill=red!30, draw=red, inner sep=2pt] (out\i) at (1.8,\y) {};
                
            \foreach \i in {1,2,3} \foreach \j in {1,2,3,4} \draw[gray!30, thin] (in\i) -- (hid\j);
            \foreach \i in {1,2,3,4} \foreach \j in {1,2,3} \draw[gray!30, thin] (hid\i) -- (out\j);
        }
    },
    pics/knn1d/.style={
        code={
            \draw[thick, gray] (0,0) -- (3.5,0); 
            \foreach \x in {0.2, 0.8, 2.6, 3.2} \fill[gray] (\x,0) circle (2pt); 
            \fill[blue] (1.65,0) circle (2.8pt) node[above, scale=0.75, blue] {$\text{SEC}(T)$}; 
            \draw[->, red, very thick] (1.6, -0.3) -- (2.6, -0.15) node[midway, below, scale=0.7] {NN}; 
            \fill[red] (2.6,0) circle (2.8pt) node[above, scale=0.75, red] {$\text{SEC}(A_{i^*})$}; 
        }
    }
}

\node[block, fill=blue!5] (input_set) {
    Calibration Images\
    $\{A_1, \dots, A_D\}$
};

\node[block, below=of input_set] (optim) {
    Grid Search for optimal $\freqParamMath_i^*$
};

\node[block, below=of optim, fill=yellow!10] (cal_db) {
    \textbf{Calibration Set}\
    $\{(\text{SEC}(A_i), \freqParamMath^*_i)\}_{i=1}^D$
};

\draw[arrow] (input_set) -- (optim);
\draw[arrow] (optim) -- (cal_db);

\node[block, fill=red!5, right=1.5cm of input_set] (target) {
    New Target\
    Image $T$
};

\node[block, below=of target] (target_sec) {
    Compute $\text{SEC}(T)$ \\
    \tikz{\pic {spectrum};}
};

\node[block, below=of target_sec] (knn) {
    Nearest Neighbor\\
    $i^* = \arg\min_i |\text{SEC}(T) - \text{SEC}(A_i)|$\\
    \tikz{\pic {knn1d};}
};

\node[block, below=of knn] (train) {
    Init with $\freqParamMath^*_{i^*}$ and Train\\
    \tikz{\pic {neuralnet};}
};

\draw[arrow] (target) -- (target_sec);
\draw[arrow] (target_sec) -- (knn);
\draw[arrow] (knn) -- (train);

\draw[arrow, dashed, red!70!black] (knn.west) -- ++(-3.,0) -| (cal_db.south) node[near end, below right, font=\scriptsize] {Lookup};

\node[above=0.2cm of input_set, font=\bfseries\color{blue!50!black}] {Offline Calibration Phase};
\node[above=0.2cm of target, font=\bfseries\color{red!50!black}] {Online Hyperparameter Selection};

\end{tikzpicture}
}
    \vspace{-7pt}
    \caption{The SEC-Conf pipeline for INR hyperparameter selection. A calibration set is used to map Spectral Energy Centroids (SEC) to optimal frequency parameters.
    For a new target image, the optimal parameter is retrieved via nearest-neighbor matching in the centroid space.}
    \label{fig:sec_conf_pipeline}
    \vspace{-15pt}
    \end{wrapfigure}
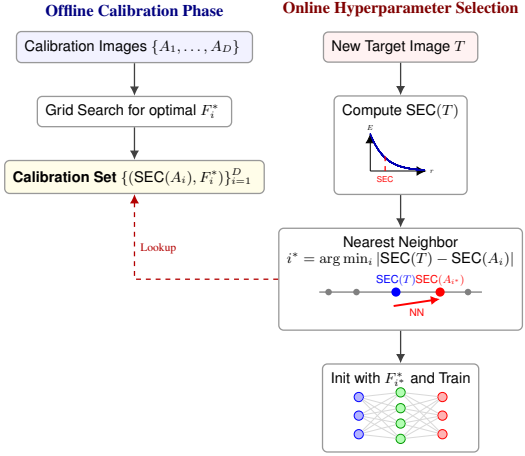

    \subsection{Spectral energy centroid}
    \textbf{\Inherentfreq{}.}
    While the energy spectrum describes the distribution of frequencies in
    detail,
    it is high-dimensional and difficult to use for direct comparison or
    optimization.
    To summarize the spectrum with a single statistic,
    we compute its mean weighted frequency.
    First, we normalize the energy spectrum to obtain a
    distribution $\normalPower(A) = \frac{\power(A)}{\|\power(A)\|_1}$.
    The \inherentfreq{} of the image is then defined as the weighted average of
    frequencies under this normalized distribution (see visualization of this
    pipeline in \cref{fig:inherent_freq_pipeline}):

    \begin{equation}
        \inherentFreqMath(A) = \sum_{r=1}^R r \normalPower(A, r),
    \end{equation}

    where $R$ is the highest radius that fits on an image.
    This quantity provides a robust measure of the central frequency with a
    clear interpretation - SEC represents the mean frequency weighted by signal power.
    We can compute $\inherentFreqMath$ for:
    1. target images to quantify their frequency content,
    2. outputs of untrained models to approximate their spectral bias.
    For a model initialized with frequency parameter $\freqParamMath$, which
    outputs image $A_{\text{init}}$ at initialization,
    we denote its \inherentfreq{} as
    $\inherentFreqMath(\text{Model}_\freqParamMath)=\inherentFreqMath(A_{\text{init}})$.
    We perform an ablation on design of \inherentfreq{} in \cref{sec:ablation_studies}.

    \textbf{Embedding configuration via \inherentfreq{}.}
    The optimal frequency parameter depends on both the target image's spectral
    properties and the model's spectral bias, both of which can be measured
    by computing \inherentfreq{} for respective signals.
    However, the measurement of spectral bias is architecture dependent and
    can only be considered an approximation.
    To address this, we propose a data-driven approach.
    We assume that images with similar \inherentfreq{} scores share similar
    optimal configurations for a given architecture.
    We utilize a small calibration set of $D$ examples $\{(A_i, \freqParamMath_i)\}_{i=1}^D$,
    where $A_i$ is an image and $\freqParamMath_i^*$ is the corresponding optimal
    \freqparam{} parameter (e.g., found through grid search).
    Calibration images are randomly selected from the dataset.
    To configure an INR model for a new test image $T$, we set the relevant \freqparam{}
    parameter to $\freqParamMath{}_{i^*}$ by finding the nearest example from the calibration set:

    \begin{equation}
        i^* = \argmin_i  |\inherentFreqMath{}(T) - \inherentFreqMath{}(A_i)|
    \end{equation}

    We refer to this approach as Spectral Energy Centroid-based Configuration
    (\our{}), and visualize the pipeline in \cref{fig:sec_conf_pipeline}.

        \begin{figure}
        \centering
        \resizebox{0.9\textwidth}{!}{
            \centering
    \begin{tikzpicture}
        \begin{groupplot}[
            group style={
                group size=4 by 2,
                horizontal sep=1.0cm,
                vertical sep=0.7cm,
            },
            width=0.27\textwidth,
            height=3.5cm,
            grid=both,
            every axis title/.style={at={(0.5,1.1)},anchor=south},
            tick label style={font=\tiny},
            label style={font=\small},
            title style={yshift=-2mm},
            x label style={yshift=1.5mm}
        ]

        \nextgroupplot[
            title={Siren},
            ylabel={\small SEC},
            xlabel={\tiny Hidden Depth},
            error bars/y dir=both,
            error bars/y explicit,
        ]  
            \addplot[dashed,mark size=1.7pt,mark=*, line width=0.3pt, color=red] table[col sep=comma, x=hidden_layers, y=center of mass_mean, y error expr=1.96*\thisrow{center of mass_sem}]{data/grouped_untrained_layers/model_siren_omega_30.0.csv};
            \addplot[dashed,mark size=1.7pt,mark=*, line width=0.3pt, color=blue] table[col sep=comma, x=hidden_layers, y=center of mass_mean, y error expr=1.96*\thisrow{center of mass_sem}]{data/grouped_untrained_layers/model_siren_omega_60.0.csv};
            \addplot[dashed,mark size=1.7pt,mark=*, line width=0.3pt, color=green] table[col sep=comma, x=hidden_layers, y=center of mass_mean, y error expr=1.96*\thisrow{center of mass_sem}]{data/grouped_untrained_layers/model_siren_omega_90.0.csv};

        \nextgroupplot[
            title={\ffeatures{}},
            xlabel={\tiny Hidden Depth},
            error bars/y dir=both,
            error bars/y explicit,
        ]
            \addplot[dashed,mark size=1.7pt,mark=*, line width=0.3pt, color=red] table[col sep=comma, x=hidden_layers, y=center of mass_mean, y error expr=1.96*\thisrow{center of mass_sem}]{data/grouped_untrained_layers/model_relu_sigma_1.0.csv};
            \addplot[dashed,mark size=1.7pt,mark=*, line width=0.3pt, color=blue] table[col sep=comma, x=hidden_layers, y=center of mass_mean, y error expr=1.96*\thisrow{center of mass_sem}]{data/grouped_untrained_layers/model_relu_sigma_3.0.csv};
            \addplot[dashed,mark size=1.7pt,mark=*, line width=0.3pt, color=green] table[col sep=comma, x=hidden_layers, y=center of mass_mean, y error expr=1.96*\thisrow{center of mass_sem}]{data/grouped_untrained_layers/model_relu_sigma_5.0.csv};

        \nextgroupplot[
            title={Finer},
            xlabel={\tiny Hidden Depth},
            error bars/y dir=both,
            error bars/y explicit,
        ]
            \addplot[dashed,mark size=1.7pt,mark=*, line width=0.3pt, color=green] table[col sep=comma, x=hidden_layers, y=center of mass_mean, y error expr=1.96*\thisrow{center of mass_sem}]{data/grouped_untrained_layers/model_finer_omega_30.0_bias_1.0.csv};
            \addplot[dashed,mark size=1.7pt,mark=*, line width=0.3pt, color=purple] table[col sep=comma, x=hidden_layers, y=center of mass_mean, y error expr=1.96*\thisrow{center of mass_sem}]{data/grouped_untrained_layers/model_finer_omega_60.0_bias_1.0.csv};
            \addplot[dashed,mark size=1.7pt,mark=*, line width=0.3pt, color=orange] table[col sep=comma, x=hidden_layers, y=center of mass_mean, y error expr=1.96*\thisrow{center of mass_sem}]{data/grouped_untrained_layers/model_finer_omega_90.0_bias_1.0.csv};

        \nextgroupplot[
            title={Wire},
            xlabel={\tiny Hidden Depth},
            error bars/y dir=both,
            error bars/y explicit,
        ]
            \addplot[dashed,mark size=1.7pt,mark=*, line width=0.3pt, color=red] table[col sep=comma, x=hidden_layers, y=center of mass_mean, y error expr=1.96*\thisrow{center of mass_sem}]{data/grouped_untrained_layers/model_wire_omega_10.0_omegai_10.0_scale_10.0.csv};
            \addplot[dashed,mark size=1.7pt,mark=*, line width=0.3pt, color=green] table[col sep=comma, x=hidden_layers, y=center of mass_mean, y error expr=1.96*\thisrow{center of mass_sem}]{data/grouped_untrained_layers/model_wire_omega_20.0_omegai_20.0_scale_10.0.csv};
            \addplot[dashed,mark size=1.7pt,mark=*, line width=0.3pt, color=magenta] table[col sep=comma, x=hidden_layers, y=center of mass_mean, y error expr=1.96*\thisrow{center of mass_sem}]{data/grouped_untrained_layers/model_wire_omega_30.0_omegai_30.0_scale_10.0.csv};

        \nextgroupplot[
            ylabel={\small SEC},
            xlabel={\tiny Hidden Width},
            xmode=log,
            error bars/y dir=both,
            error bars/y explicit,
            legend style={at={(0.5,-0.35)},anchor=north, font=\tiny},
            legend columns=1,
        ]
            \addplot[dashed,mark size=1.7pt,mark=*, line width=0.3pt, color=red] table[col sep=comma, x=hidden_dim, y=center of mass_mean, y error expr=1.96*\thisrow{center of mass_sem}]{data/grouped_untrained_dim/model_siren_omega_30.0.csv};
            \addlegendentry{\tiny$\omega_s=30.0$}
            \addplot[dashed,mark size=1.7pt,mark=*, line width=0.3pt, color=blue] table[col sep=comma, x=hidden_dim, y=center of mass_mean, y error expr=1.96*\thisrow{center of mass_sem}]{data/grouped_untrained_dim/model_siren_omega_60.0.csv};
            \addlegendentry{\tiny$\omega_s=60.0$}
            \addplot[dashed,mark size=1.7pt,mark=*, line width=0.3pt, color=green] table[col sep=comma, x=hidden_dim, y=center of mass_mean, y error expr=1.96*\thisrow{center of mass_sem}]{data/grouped_untrained_dim/model_siren_omega_90.0.csv};
            \addlegendentry{\tiny$\omega_s=90.0$}

        \nextgroupplot[
            xlabel={\tiny Hidden Width},
            xmode=log,
            error bars/y dir=both,
            error bars/y explicit,
            legend style={at={(0.5,-0.35)},anchor=north, font=\tiny},
            legend columns=1,
        ]
            \addplot[dashed,mark size=1.7pt,mark=*, line width=0.3pt, color=red] table[col sep=comma, x=hidden_dim, y=center of mass_mean, y error expr=1.96*\thisrow{center of mass_sem}]{data/grouped_untrained_dim/model_relu_sigma_1.0.csv};
            \addlegendentry{\tiny$\sigma=1.0$}
            \addplot[dashed,mark size=1.7pt,mark=*, line width=0.3pt, color=blue] table[col sep=comma, x=hidden_dim, y=center of mass_mean, y error expr=1.96*\thisrow{center of mass_sem}]{data/grouped_untrained_dim/model_relu_sigma_3.0.csv};
            \addlegendentry{\tiny$\sigma=3.0$}
            \addplot[dashed,mark size=1.7pt,mark=*, line width=0.3pt, color=green] table[col sep=comma, x=hidden_dim, y=center of mass_mean, y error expr=1.96*\thisrow{center of mass_sem}]{data/grouped_untrained_dim/model_relu_sigma_5.0.csv};
            \addlegendentry{\tiny$\sigma=5.0$}

        \nextgroupplot[
            xlabel={\tiny Hidden Width},
            xmode=log,
            error bars/y dir=both,
            error bars/y explicit,
            legend style={at={(0.5,-0.35)},anchor=north, font=\tiny},
            legend columns=1,
        ]
            \addplot[dashed,mark size=1.7pt,mark=*, line width=0.3pt, color=green] table[col sep=comma, x=hidden_dim, y=center of mass_mean, y error expr=1.96*\thisrow{center of mass_sem}]{data/grouped_untrained_dim/model_finer_omega_30.0_bias_1.0.csv};
            \addlegendentry{\tiny$\omega_f=30.0$}
            \addplot[dashed,mark size=1.7pt,mark=*, line width=0.3pt, color=purple] table[col sep=comma, x=hidden_dim, y=center of mass_mean, y error expr=1.96*\thisrow{center of mass_sem}]{data/grouped_untrained_dim/model_finer_omega_60.0_bias_1.0.csv};
            \addlegendentry{\tiny$\omega_f=60.0$}
            \addplot[dashed,mark size=1.7pt,mark=*, line width=0.3pt, color=orange] table[col sep=comma, x=hidden_dim, y=center of mass_mean, y error expr=1.96*\thisrow{center of mass_sem}]{data/grouped_untrained_dim/model_finer_omega_90.0_bias_1.0.csv};
            \addlegendentry{\tiny$\omega_f=90.0$}

        \nextgroupplot[
            xlabel={\tiny Hidden Width},
            xmode=log,
            error bars/y dir=both,
            error bars/y explicit,
            legend style={at={(0.5,-0.35)},anchor=north, font=\tiny},
            legend columns=1,
        ]
            \addplot[dashed,mark size=1.7pt,mark=*, line width=0.3pt, color=red] table[col sep=comma, x=hidden_dim, y=center of mass_mean, y error expr=1.96*\thisrow{center of mass_sem}]{data/grouped_untrained_dim/model_wire_omega_10.0_omegai_10.0_scale_10.0.csv};
            \addlegendentry{\tiny$\omega_w=10.0$}
            \addplot[dashed,mark size=1.7pt,mark=*, line width=0.3pt, color=green] table[col sep=comma, x=hidden_dim, y=center of mass_mean, y error expr=1.96*\thisrow{center of mass_sem}]{data/grouped_untrained_dim/model_wire_omega_20.0_omegai_20.0_scale_10.0.csv};
            \addlegendentry{\tiny$\omega_w=20.0$}
            \addplot[dashed,mark size=1.7pt,mark=*, line width=0.3pt, color=magenta] table[col sep=comma, x=hidden_dim, y=center of mass_mean, y error expr=1.96*\thisrow{center of mass_sem}]{data/grouped_untrained_dim/model_wire_omega_30.0_omegai_30.0_scale_10.0.csv};
            \addlegendentry{\tiny$\omega_w=30.0$}

        \end{groupplot}
    \end{tikzpicture}
        }
        \caption{Relation between \Inherentfreq{} and model size.
        We measure \inherentfreq{} across various model sizes, by adjusting the
        number of layers or width of a 3-layer, 256-width MLP.
        We note that \inherentfreq{} can significantly increase with depth of
        the model, explaining why \fresh{} tends to incorrectly select lower
        \freqparam{} parameter values for deeper models.
        However, \inherentfreq{} is not particularly affected by model width
        (except for \wire{}), suggesting that increasing width could be a safer
        strategy for scaling model capacity, as it preserves spectral properties.
        We note that \wire{} is a significant outlier, with extremely high
        frequencies present across almost all model sizes - we believe this
        high spectral bias explains why this architecture tends to perform
        poorly.
        \inherentfreq{} was measured 10 times at each frequency, and we plot
        both mean and 95\% confidence interval.
        }
        \label{fig:freq_vs_width_and_layers}
       \vspace{-10pt}
    \end{figure}
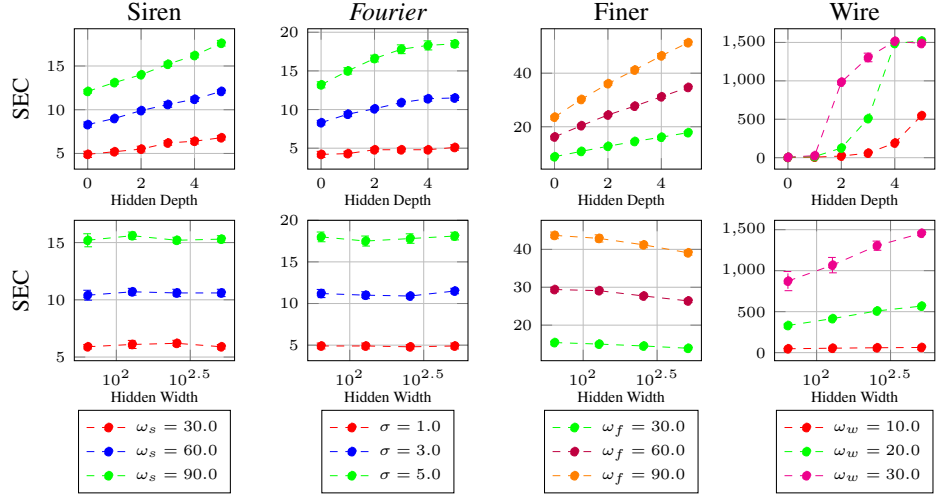

    \textbf{Frequency matching.}
    Transferring optimal \freqparam{} parameter configurations between
    different INR architectures has the potential to make model comparisons
    fairer by aligning their spectral biases, allowing to attribute any
    changes in observed performance strictly to architectural differences.
    Furthermore, it could simplify parameter selection for new models
    by leveraging well-configured models (e.g., \finer{}) to
    select the initial set of tested parameters.
    However, transferring frequency parameters between models is challenging
    as frequency parameters (e.g., $\omega_0$ in \siren{}, $\sigma$ in
    \ffeatures{}) have different interpretations and effects on the model.
    Consequently, hyperparameter values cannot be directly reused between them.

    We hypothesize that while specific parameter values are not portable, the
    optimal spectral bias is at least partially transferable.
    Therefore, we propose Frequency Matching: a method to configure a target
    model by aligning its spectral bias with that of a reference model.
    Given a reference initialized model (e.g., a well-tuned architecture like
    \finer{}) with parameter $\freqParamMath_{ref}$,
    we denote the matched parameter value for a target model
    as $\operatorname{Fit}_{\freqParamMath_{ref}}(\freqParamMath)$,
    where $\freqParamMath$ is the parameter of the matched model
    (e.g., we will use $\omega_s$ when adjusting spectral bias of \siren{}).
    This value is obtained by finding the parameter $\freqParamMath$
    that minimizes the difference in spectral centroids:
        \begin{equation*}
            \operatorname{Fit}_{\freqParamMath_{ref}}(\freqParamMath) = \argmin_\freqParamMath
            |\inherentFreqMath(\text{Model}_\freqParamMath) -
            \inherentFreqMath(\text{Ref}_{\freqParamMath_{ref}})|.
        \end{equation*}

    This approach allows us to translate configurations across disparate
    architectures without requiring expensive re-optimization.

    \section{Experiments}
    \label{sec:experiments}

    \Inherentfreq{} serves as a compact and interpretable summary of frequency
    content.
    In this section, we first analyze how model architecture influences
    \inherentfreq{}, providing insights into spectral bias.
    Building on this, we validate \inherentfreq's practical utility by applying it to three
    distinct problems:
    guiding the selection of optimal \freqparam{} parameters for new images,
    analyzing the relationship between image complexity and reconstruction
    quality,
    and aligning the frequency profiles of different architectures.

    We conduct our experiments on a diverse benchmark of 100 images sampled
    from the LIU4K-v2 dataset \cite{Liu4K}, all down-sampled to a resolution of
    $2\text{k} \times 1.4\text{k}$.
    To evaluate our configuration strategies, we employ a random 10/90 calibration/test
    split, where 10 images are used to calibrate \our{} and
    the remaining 90 images serve as a
    held-out test set.
    We evaluate performance across three distinct model scales: small (S,
    $2\times128$), medium (M, $3\times256$), and large (L, $4\times512$).
    Our benchmark includes four representative INR architectures: \siren{},
    \wire{}, \ffeatures{}, and \finer{}.
    For each architecture and model size, we perform an exhaustive grid search
    over their primary \freqparam{} parameters:
    $\omega_s \in \{30, 40, \dots, 110\}$,
    $\omega_f \in \{10, 20, \dots, 100\}$,
    $\omega_w \in \{1, 5, 10, \dots, 30\}$, and
    $\sigma \in \{1, 2, \dots, 9\}$.
    All models are trained over 3 seeds for 15,000 optimization steps using the Adam
    optimizer \cite{kingma2014adam} in PyTorch \cite{paszke2019pytorch} 
    We describe compute resources used in \cref{sec:compute_resources}.

    \begin{wraptable}{r}{0.55\textwidth}
        \centering
       \vspace{-13pt}
        \caption{Comparison of reconstruction quality (PSNR) for models
        configured via our method versus baseline defaults and \fresh{}.
                We report mean performance and standard error computed across 3 seeds.
        }
        \vspace{3pt}
        \begin{footnotesize}
            \begin{sc}
                \begin{tabular}{lccc}
\toprule
psnr $\uparrow$ & S & M & L \\
\midrule
\siren{} & $26.03$ $\mathsmaller{\pm 0.01}$ & $29.97$ $\mathsmaller{\pm 0.01}$ & $32.06$ $\mathsmaller{\pm 0.10}$ \\
+best & $26.64$ $\mathsmaller{\pm 0.01}$ & $31.31$ $\mathsmaller{\pm 0.01}$ & $34.60$ $\mathsmaller{\pm 0.03}$ \\
+\fresh{} & $26.45$ $\mathsmaller{\pm 0.02}$ & $31.21$ $\mathsmaller{\pm 0.01}$ & $34.07$ $\mathsmaller{\pm 0.07}$ \\
+\our{} & $\mathbf{26.61}$ $\mathsmaller{\pm 0.01}$ & $\mathbf{31.25}$ $\mathsmaller{\pm 0.02}$ & $\mathbf{34.49}$ $\mathsmaller{\pm 0.05}$ \\
\midrule
\ffeatures{} & $24.72$ $\mathsmaller{\pm 0.04}$ & $27.68$ $\mathsmaller{\pm 0.11}$ & $29.01$ $\mathsmaller{\pm 0.03}$ \\
+best & $26.54$ $\mathsmaller{\pm 0.05}$ & $30.75$ $\mathsmaller{\pm 0.05}$ & $32.14$ $\mathsmaller{\pm 0.06}$ \\
+\fresh{} & $\mathbf{26.35}$ $\mathsmaller{\pm 0.05}$ & $30.05$ $\mathsmaller{\pm 0.06}$ & $30.92$ $\mathsmaller{\pm 0.02}$ \\
+\our{} & $\mathbf{26.35}$ $\mathsmaller{\pm 0.07}$ & $\mathbf{30.45}$ $\mathsmaller{\pm 0.14}$ & $\mathbf{31.74}$ $\mathsmaller{\pm 0.08}$ \\
\midrule
\finer{} & $\mathbf{27.26}$ $\mathsmaller{\pm 0.02}$ & $31.86$ $\mathsmaller{\pm 0.01}$ & $34.23$ $\mathsmaller{\pm 0.01}$ \\
+best & $27.32$ $\mathsmaller{\pm 0.01}$ & $32.13$ $\mathsmaller{\pm 0.03}$ & $36.00$ $\mathsmaller{\pm 0.11}$ \\
+\fresh{} & $27.15$ $\mathsmaller{\pm 0.02}$ & $\mathbf{32.01}$ $\mathsmaller{\pm 0.01}$ & $34.91$ $\mathsmaller{\pm 0.02}$ \\
+\our{} & $27.21$ $\mathsmaller{\pm 0.02}$ & $31.99$ $\mathsmaller{\pm 0.01}$ & $\mathbf{35.81}$ $\mathsmaller{\pm 0.07}$ \\
\midrule
\wire{} & $24.39$ $\mathsmaller{\pm 0.02}$ & $27.70$ $\mathsmaller{\pm 0.04}$ & $28.67$ $\mathsmaller{\pm 0.08}$ \\
+best & $26.10$ $\mathsmaller{\pm 0.02}$ & $29.47$ $\mathsmaller{\pm 0.02}$ & $31.76$ $\mathsmaller{\pm 0.09}$ \\
+\fresh{} & n/a & n/a & n/a \\
+\our{} & $\mathbf{26.07}$ $\mathsmaller{\pm 0.01}$ & $\mathbf{29.43}$ $\mathsmaller{\pm 0.03}$ & $\mathbf{31.58}$ $\mathsmaller{\pm 0.17}$ \\
\bottomrule
\end{tabular}

            \end{sc}
        \end{footnotesize}
        \label{tab:freq_param_prediction_psnr}
       \vspace{-5pt}
    \end{wraptable}

    \subsection{Effects of architecture
    }
    \label{sec:effects_of_architecture_on_sec}

    We start by analyzing how the spectral bias of models, as approximated by
    \inherentfreq{}, varies with model architecture and size, with results
    presented in \cref{fig:freq_vs_width_and_layers}.
    Our analysis reveals that \inherentfreq{} increases significantly with
    model depth across almost all architectures,
    confirming empirically theoretical results from previous work
    \cite{inr_dictionaries2022}.
    This likely explains the behavior of \fresh{} in \cref{fig:sigma_vs_model_size}
    - the method accounts for the depth-related increase in frequency by
    lowering the value of \freqparam{}.
    This depth-induced increase in \inherentfreq{} shows that deeper
    models naturally favor higher frequencies and may require lower 
    embedding parameters to achieve
    the same initial spectral bias as shallower counterparts.
    However, our results indicate that the increased capacity of deeper
    models allows them to better utilize high-frequency inputs.
    In our experiments, this capacity effect dominates, and deeper models
    consistently benefit from higher embedding frequencies.

    \begin{wraptable}{r}{0.45\textwidth}
        \centering
       \vspace{-13pt}
        \caption{Accuracy of \fresh{} and \our{} in predicting the optimal frequency parameter.
        While \our{}'s accuracy is relatively low, just approximately tracking the oracle 
        (see \cref{fig:sigma_vs_model_size}) is enough for good performance.
        }
        \vspace{2pt}
        \begin{small}
            \begin{sc}
                \begin{tabular}{lccc}
\toprule
Accuracy $\uparrow$ & L & M & S \\
\midrule
\siren{} + \fresh{} & $0.16$  & $0.17$  & $0.01$  \\
\siren{} + \our{} & $0.64$  & $0.21$  & $0.36$  \\
\midrule
\ffeatures{} + \fresh{} & $0.01$  & $0.01$  & $0.18$  \\
\ffeatures{} + \our{} & $0.32$  & $0.20$  & $0.21$  \\
\midrule
\finer{} + \fresh{} & $0.0$ & $0.20$  & $0.06$  \\
\finer{} + \our{} & $0.49$  & $0.18$  & $0.22$  \\
\midrule
\wire{} + \fresh{} & n/a & n/a & n/a \\
\wire{} + \our{} & $0.47$  & $0.47$  & $0.57$  \\
\bottomrule
\end{tabular}

            \end{sc}
        \end{small}
        \label{tab:freq_param_prediction_accuracy}
    \end{wraptable}
    
    In contrast to depth, model width has a negligible effect on spectral
    bias for most architectures.
    This suggests that scaling a model by width is more predictable than
    scaling by depth in terms of frequency characteristics.
    However, there are notable exceptions.
    For instance, the \wire{} architecture stands as an outlier, with a
    significantly higher \inherentfreq{} that shows a unique sensitivity
    to changes in width.
    We also observe that for \finer{}, increased width correlates with
    a decrease in \inherentfreq{}.
    This phenomenon is caused by the use of \siren{} weight
    initialization, which was tailored to sinusoidal (not variable-periodic)
    activations and assumes that inputs to each layer follow a
    standard normal distribution
    (see further discussion in \cref{sec:finer_init_issue}).
    Such complex interactions between architecture and spectral bias underscore
    the necessity of a measurement-based approach like \inherentfreq{}
    for robust model configuration.

    \subsection{Prediction of embedding frequency parameters}
    \label{sec:inh_freq_pred}

    As discussed in \cref{sec:motivation}, the optimal
    frequency parameter depends on both the target image's spectral properties
    and the model's capacity.
    Existing heuristics like \fresh{} account only for differences stemming
    from the target image and fail to account for architectural factors.
    In this section, we evaluate the performance of the \our{}-based
    configuration method (described in \cref{sec:method}), which leverages a
    small calibration set to predict optimal \freqparam{} parameters.

    \Cref{tab:freq_param_prediction_psnr} compares our method against \fresh{}
    and the standard baseline of default parameters.
    Our approach consistently outperforms the baseline and matches or exceeds
    the performance of \fresh{}, primarily because it accounts for
    architectural bias.
    This advantage is reflected in both the final reconstruction quality and
    the selection accuracy
    (see \cref{tab:freq_param_prediction_psnr,tab:freq_param_prediction_accuracy}).
    We note that the absolute PSNR gap
    between \our{} and \fresh{} is bounded by
    the difference between the oracle and baseline configuration.
    In settings where the default is already
    near-optimal, no method - including ours - can show
    large gains. Where the gap is larger, \our{}
    reliably closes it, e.g. for large models.
    For this reason, we also measure selection accuracy
    \cref{tab:freq_param_prediction_accuracy}, which
    better shows that \our{} demonstrates a substantially
    higher hit rate in identifying the optimal configuration
    as compared to the \fresh{} heuristic.
    The performance gain is particularly pronounced for larger models, where
    standard conservative defaults underestimate the required frequency
    and \fresh{} fails to account for increased model size
    (as discussed in \cref{sec:motivation}).
    While \fresh{} tends to suggest lower frequency parameters for larger
    models, our data-driven approach identifies that these models
    require higher frequency parameters.
    Additionally, we observe that while \wire{} possesses a significantly
    higher \inherentfreq{} than other architectures, this does not translate to
    superior performance.
    As shown in \cref{tab:freq_param_prediction_psnr}, \siren{} consistently
    outperforms \wire{} across all model sizes, confirming that excessive
    high-frequency bias is detrimental.
    Additionally, we provide detailed training curves
    in \cref{subsec:additional_sec_conf}, a cross-domain generalization study in \cref{sec:cross_domain} and additional metrics in \cref{sec:additional_metrics}.

    Lastly, we acknowledge that while preparing the calibration dataset requires an upfront cost, results
    from our experiment can be reused for new images.
    We publish them together with a script to predict optimal \freqparam{}
    parameter values for new target signals.
\footnote{https://github.com/gmum/SEC}

    \begin{figure}[ht]
        \centering
        \begin{minipage}{0.61\textwidth}
            \centering
                \input{experiment_results/psnr_vs_freq_statistic}
            \caption{Relationship between \inherentfreq{} and reconstruction quality for Medium-sized models using default \freqparam{} values. 
            There is a strong correlation between \inherentfreq{} and PSNR, reflecting that \inherentfreq{} captures images complexity well.
            }
            \label{fig:psnr_from_frequency}
        \end{minipage}\hfill
        \begin{minipage}{0.36\textwidth}
            \centering
            \captionof{table}{Spearman’s rank correlation between \inherentfreq{} and reconstruction quality (PSNR) for baseline models of size M.}
            \label{tab:psnr_correlation}
            \begin{footnotesize}
                \begin{sc}
                    \begin{tabular}{lccc}
\toprule
Corr $\downarrow$ & S & M & L \\
\midrule
siren & -0.866 & -0.822 & -0.760 \\
relu & -0.839 & -0.834 & -0.845 \\
finer & -0.861 & -0.796 & -0.739 \\
wire & -0.849 & -0.843 & -0.695 \\
\bottomrule
\end{tabular}

                \end{sc}
            \end{footnotesize}
        \end{minipage}
        \vspace{-5pt}
    \end{figure}

    \subsection{Spectral energy centroid and image complexity}

    Beyond hyperparameter selection, \inherentfreq{} serves as a robust proxy
    for image complexity, which we validate in this section
    by showing how increasing \inherentfreq{} correlates with
    a decreasing reconstruction quality across standard INR architectures.
    We present this relationship in \cref{fig:psnr_from_frequency} for our 100-image
    dataset.
    In \cref{tab:psnr_correlation} we observe a strongly negative 
    Spearman’s rank correlation across all architectures.
    This confirms that images with higher spectral centroids - corresponding to
    finer details and textures - are harder to represent, resulting in lower PSNR
    values.
    While the correlation is significant, the spread in the data suggests that
    \inherentfreq{} alone does not fully dictate learnability. However, it
    remains a dominant factor.
    This relationship is additionally confirmed through images in \cref{fig:inherent_freq_pipeline}
    and model outputs in \cref{sec:sec_visualisation_init_extras}.
    This analysis validates our score as a meaningful, scalar summary of signal
    difficulty, but also highlights the need for
    further investigation.

    \subsection{INR alignment via frequency matching}
    \label{sec:network_matching}

    Directly comparing different INR architectures is challenging because each
    one starts with a different default frequency bias.
    To enable a fair comparison, we employ our frequency matching strategy to
    align spectral biases between architectures and isolate the effects of
    model architecture on performance.
    This ensures that performance gains are attributed to architectural
    design rather than superior tuning.
    We perform matching over the grid-search defined at the start of
    \cref{sec:experiments}.
    For robustness, the \inherentfreq{} values used for matching are
    calculated as the mean over 10 seeds.
    The results of this alignment are summarized in
    \cref{tab:network_matching}, where we treat each architecture as a
    reference and match all other models to its initial spectral bias.
    See \cref{sec:matching_extras} for a discussion of specific parameter values
    used.

    We first observe that standard \siren{} and \ffeatures{} models are
    typically initialized with similar, relatively low frequency biases.
    Consequently, matching between them results in no parameter changes
    and maintains their baseline performance gaps. However,
    when we align these conservative architectures with the \finer{}
    model, both \siren{} and \ffeatures{} see significant performance gains.
    This indicates that their default configurations are overly
    conservative, preventing them from capturing fine details.
    \begin{wraptable}{r}{0.58\textwidth}
        \centering
        \caption{
            Comparison of reconstruction quality for baseline models
            versus models with frequency parameters matched to a reference
            architecture.
            Note that for \siren{} and \ffeatures{}, the spectral profiles are
            similar, resulting in no parameter change.
            We exclude results for matching to \wire{} M and L models as their
            exceptionally high frequencies could not be matched within our
            hyperparameter grid.
            Results are averaged over 3 seeds, and we report standard error.
        }
        \vspace{-2pt}
        \begin{center}
            \begin{footnotesize}
                \begin{sc}
                    \begin{tabular}{lccc}
\toprule
PSNR $\uparrow$ & S & M & L \\
\midrule
Ref: \finer & $\mathbf{27.09}$ $\mathsmaller{\pm 0.02}$ & $\mathbf{31.75}$ $\mathsmaller{\pm 0.02}$ & $34.14$ $\mathsmaller{\pm 0.01}$ \\
$\operatorname{Fit}_{\omega_f=30}(\sigma)$ & $26.13$ $\mathsmaller{\pm 0.09}$ & $29.95$ $\mathsmaller{\pm 0.05}$ & $30.72$ $\mathsmaller{\pm 0.03}$ \\
$\operatorname{Fit}_{\omega_f=30}(\omega_s)$ & $26.37$ $\mathsmaller{\pm 0.02}$ & $31.13$ $\mathsmaller{\pm 0.01}$ & $\mathbf{34.21}$ $\mathsmaller{\pm 0.05}$ \\
$\operatorname{Fit}_{\omega_f=30}(\omega_w)$ & $25.90$ $\mathsmaller{\pm 0.02}$ & --- & --- \\
\midrule
Ref: \wire & $25.67$ $\mathsmaller{\pm 0.01}$ & $28.84$ $\mathsmaller{\pm 0.03}$ & $31.18$ $\mathsmaller{\pm 0.12}$ \\
$\operatorname{Fit}_{\omega_w=10}(\omega_f)$ & $\mathbf{26.98}$ $\mathsmaller{\pm 0.02}$ & --- & --- \\
$\operatorname{Fit}_{\omega_w=10}(\sigma)$ & $26.17$ $\mathsmaller{\pm 0.03}$ & --- & --- \\
$\operatorname{Fit}_{\omega_w=10}(\omega_s)$ & $26.12$ $\mathsmaller{\pm 0.03}$ & --- & --- \\
\midrule
Ref: \siren & $25.86$ $\mathsmaller{\pm 0.01}$ & $29.84$ $\mathsmaller{\pm 0.01}$ & $\mathbf{31.97}$ $\mathsmaller{\pm 0.09}$ \\
$\operatorname{Fit}_{\omega_s=30}(\omega_f)$ & $\mathbf{26.39}$ $\mathsmaller{\pm 0.04}$ & $\mathbf{30.16}$ $\mathsmaller{\pm 0.02}$ & $31.81$ $\mathsmaller{\pm 0.03}$ \\
$\operatorname{Fit}_{\omega_s=30}(\sigma)$ & $24.52$ $\mathsmaller{\pm 0.05}$ & $27.53$ $\mathsmaller{\pm 0.09}$ & $28.87$ $\mathsmaller{\pm 0.03}$ \\
$\operatorname{Fit}_{\omega_s=30}(\omega_w)$ & $25.90$ $\mathsmaller{\pm 0.02}$ & --- & --- \\
\midrule
Ref: \ffeatures & $24.52$ $\mathsmaller{\pm 0.05}$ & $27.53$ $\mathsmaller{\pm 0.09}$ & $28.87$ $\mathsmaller{\pm 0.03}$ \\
$\operatorname{Fit}_{\sigma=1.0}(\omega_f)$ & $\mathbf{26.39}$ $\mathsmaller{\pm 0.04}$ & $\mathbf{30.16}$ $\mathsmaller{\pm 0.02}$ & $31.81$ $\mathsmaller{\pm 0.03}$ \\
$\operatorname{Fit}_{\sigma=1.0}(\omega_s)$ & $25.86$ $\mathsmaller{\pm 0.01}$ & $29.84$ $\mathsmaller{\pm 0.01}$ & $\mathbf{31.97}$ $\mathsmaller{\pm 0.09}$ \\
$\operatorname{Fit}_{\sigma=1.0}(\omega_w)$ & $25.90$ $\mathsmaller{\pm 0.02}$ & --- & --- \\
\bottomrule
\vspace{-0.7cm}
\end{tabular}

                \end{sc}
            \end{footnotesize}
        \end{center}
        \label{tab:network_matching}
        \vspace{-10pt}
    \end{wraptable}
    Notably, the large (L) \siren{} model, when matched to \finer{},
    achieves performance parity with it.
    For smaller models, while matching improves performance, it does not
    fully close the gap, suggesting that \finer{} benefits from
    architectural advantages beyond its initial spectral bias.

    As noted in \cref{sec:inh_freq_pred}, the \wire{} architecture
    presents a unique challenge due to its high default \inherentfreq{}
    and we did not find a suitable matching of \wire{} to other architectures
    for medium and large models.
    When matching small \wire{} to other architectures, we select the minimum
    available frequency parameters - we note this improves the performance of \wire{},
    which implies performance of this architecture could be further improved,
    if it was configured with a more conservative spectral bias,
    perhaps by avoiding the frequency increase at every layer.
    Conversely, matching other architectures to \wire{} often requires
    parameters beyond the searched grid.

    Despite frequency matching offering only an approximate spectral-bias alignment,
    it can significantly impact the relative performance of diverse architectures.
    We hope it can serve researchers in finding good model configurations
    both in the context of testing old architectures, and developing new ones.

    \section{Conclusion}

    In this work, we introduced \inherentfreq{} as a robust and interpretable
    metric for quantifying the frequency content of images and the spectral
    bias of INR models.
    Using this measure, we revealed that model architecture - specifically
    depth - significantly alters spectral bias, a factor overlooked by existing
    heuristics for hyperparameter configuration.
    While we demonstrated the practical utility of \inherentfreq{} through a
    data-driven hyperparameter configuration strategy and cross-architecture
    alignment,
    its primary value lies in offering a clearer lens for analyzing INR
    behavior.
    \textbf{By providing a scalar summary of spectral properties, \inherentfreq{}
    facilitates a deeper understanding of the relationship between signal
    complexity, model capacity, and reconstruction quality}.
    Our findings highlight the importance of architecture-aware frequency
    alignment and establish \inherentfreq{} as a valuable tool for future
    research.

    \section*{Limitations}
    SEC is a one-dimensional scalar summary of the
    frequency spectrum. As such, signals with different
    spectral shapes may map to the same centroid value,
    and SEC cannot capture all nuances of frequency
    distributions. 
    Additionally, SEC-conf relies on a calibration set
    obtained through offline grid search, unlike FreSh,
    which is a data-free heuristic operating on a single
    target image. This means the comparison between the
    two methods is not entirely apples-to-apples: SEC-conf
    has access to more information. This trade-off is a
    direct consequence of our goal of generality - the
    calibration set is what enables SEC to capture the
    relationship between spectral properties and model
    behavior across architectures and depths.

    {
    \small
    \bibliographystyle{plain}
    \bibliography{example_paper}
    }


\appendix

\newpage
\section{INR Architectures}
    \label{app:architectures}
    This section provides the exact mathematical formulations for the
    coordinate embeddings and activations used in the architectures
    discussed in the main text.

    \textbf{\siren{}} uses the following sinusoidal embedding:
    \begin{equation}
 \label{eq:siren_embedding}
 \gamma_S(\mathbf{x}) = \sin(\omega_s \mathbf{W} \mathbf{x} + \mathbf{b}).
    \end{equation}

    \textbf{\ffeatures{}} uses a fixed Fourier embedding:
    \begin{equation}
 \label{eq:fourier_features}
 \gamma_F(\mathbf{x}) = [\sin(2\pi \mathbf{W}\mathbf{x}), \cos(2\pi
  \mathbf{W}\mathbf{x})].
    \end{equation}

    \textbf{\wire{}} uses a complex Gabor wavelet activation:
    \begin{equation}
 \label{eq:wire_activation}
 \psi(x)=e^{j\omega_{w}x}e^{-|s_{0}x|^{2}}.
    \end{equation}

    \textbf{\finer{}} uses a variable-periodic activation:
    \begin{equation}
 \label{eq:finer_layer}
 \gamma_F(\mathbf{x}) = \sin\Bigl(\omega_f (|\mathbf{W} \mathbf{x} +
  \mathbf{b}| + 1)(\mathbf{W} \mathbf{x} +
  \mathbf{b})\Bigr).
    \end{equation}

\section{Additional results}

\subsection{Compute resources}
\label{sec:compute_resources}

We run our experiments using nodes of mixed GPUs, mainly with NVIDIA GeForce RTX 2070 and A100 40GB GPUs, with a single INR training taking up to 2 hours. In total, the grid search described in \cref{sec:experiments} took around 5'400 hours. During initial exploration of SEC, we ran additional experiments that required a fraction of compute needed for the main grid search. 

\subsection{SEC at initialization}
\label{sec:sec_visualisation_init_extras}

\begin{figure}[ht]
    \centering
     \resizebox{1.0\textwidth}{!}{
         \input{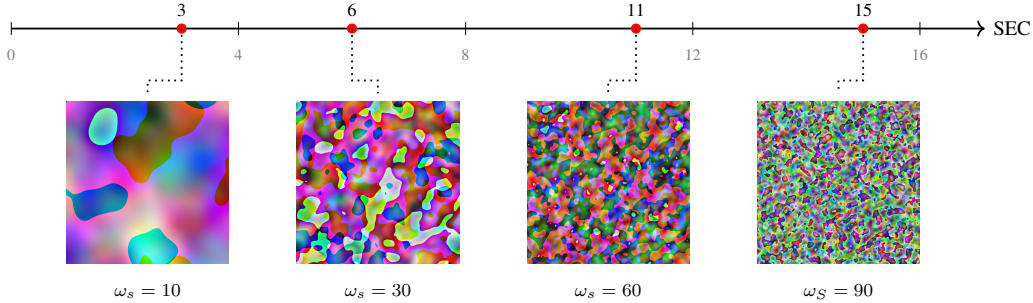}
        }
    
    \caption{We visualize the effect of the initialization parameter $\omega_s$ on the SEC of a \siren{} model.
    Higher values of $\omega_s$ result in a monotonic increase in SEC,
 corresponding to the appearance of finer spatial details in the network's initial output.}
    \label{fig:untrained_network_and_sec}
\end{figure}

We visualize the relationship between the initialization hyperparameter
$\omega_s$ and the Spectral Energy Centroid (SEC) for a \siren{} model
in \cref{fig:untrained_network_and_sec}.
As $\omega_s$ increases, the SEC shifts toward higher values.
This shift accurately reflects changes in the model's spectral bias,
where a larger SEC corresponds to finer, higher-frequency spatial features.

\subsection{Training dynamics of SEC-conf}
\label{subsec:additional_sec_conf}

\begin{figure}
    \centering

    \resizebox{0.9\textwidth}{!}{
            \input{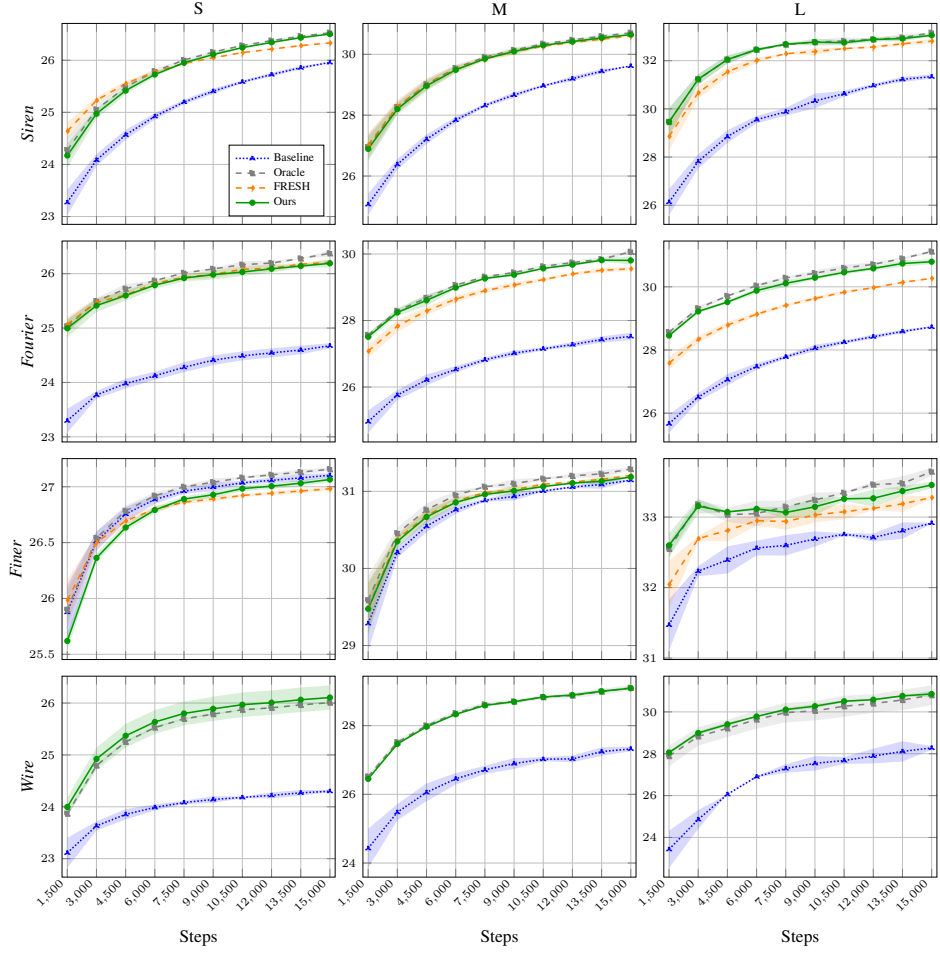}
        }

    \caption{Mean PSNR over training steps for each
    configuration method across architectures and model
    sizes. Shaded areas indicate standard error.}
    \label{fig:psnr_over_steps}
    \vskip -3mm
\end{figure}

\Cref{fig:psnr_over_steps} shows how mean PSNR evolves
over the course of training for each configuration
method. We train all models until convergence, as
suboptimal frequency configurations can temporarily
outperform better ones during early training stages.
While this effect may not always be apparent when
averaging over many images, it is visible in the 
small \siren{} setting, where \fresh{}
briefly exceeds even the oracle configuration before
falling behind.

\begin{table}[h!]
    \centering
    \caption{Comparison of reconstruction quality (SSIM) for models configured using various approaches.}
    \vspace{5pt}
    \label{tab:freq_param_prediction_ssim}
    \begin{small}
 \begin{sc}
    \begin{tabular}{lccc}
\toprule
ssim $\uparrow$ & S & M & L \\
\midrule
\siren{} & $0.680$ $\mathsmaller{\pm 0.0001}$ & $0.810$ $\mathsmaller{\pm 0.0005}$ & $0.870$ $\mathsmaller{\pm 0.0027}$ \\
+best & $0.693$ $\mathsmaller{\pm 0.0006}$ & $0.835$ $\mathsmaller{\pm 0.0002}$ & $0.900$ $\mathsmaller{\pm 0.0001}$ \\
+\fresh{} & $0.678$ $\mathsmaller{\pm 0.0011}$ & $0.828$ $\mathsmaller{\pm 0.0002}$ & $0.897$ $\mathsmaller{\pm 0.0005}$ \\
+\our{} & $\mathbf{0.692}$ $\mathsmaller{\pm 0.0006}$ & $\mathbf{0.832}$ $\mathsmaller{\pm 0.0002}$ & $\mathbf{0.899}$ $\mathsmaller{\pm 0.0004}$ \\
\midrule
\ffeatures{} & $0.651$ $\mathsmaller{\pm 0.0016}$ & $0.746$ $\mathsmaller{\pm 0.0026}$ & $0.788$ $\mathsmaller{\pm 0.0009}$ \\
+best & $0.676$ $\mathsmaller{\pm 0.0012}$ & $0.819$ $\mathsmaller{\pm 0.0006}$ & $0.863$ $\mathsmaller{\pm 0.0012}$ \\
+\fresh{} & $\mathbf{0.678}$ $\mathsmaller{\pm 0.0010}$ & $0.812$ $\mathsmaller{\pm 0.0010}$ & $0.841$ $\mathsmaller{\pm 0.0009}$ \\
+\our{} & $0.669$ $\mathsmaller{\pm 0.0020}$ & $\mathbf{0.815}$ $\mathsmaller{\pm 0.0021}$ & $\mathbf{0.858}$ $\mathsmaller{\pm 0.0035}$ \\
\midrule
\finer{} & $\mathbf{0.702}$ $\mathsmaller{\pm 0.0003}$ & $\mathbf{0.846}$ $\mathsmaller{\pm 0.0001}$ & $0.905$ $\mathsmaller{\pm 0.0002}$ \\
+best & $0.702$ $\mathsmaller{\pm 0.0003}$ & $0.843$ $\mathsmaller{\pm 0.0006}$ & $0.922$ $\mathsmaller{\pm 0.0008}$ \\
+\fresh{} & $0.684$ $\mathsmaller{\pm 0.0012}$ & $0.841$ $\mathsmaller{\pm 0.0003}$ & $0.913$ $\mathsmaller{\pm 0.0003}$ \\
+\our{} & $\mathbf{0.702}$ $\mathsmaller{\pm 0.0004}$ & $0.841$ $\mathsmaller{\pm 0.0002}$ & $\mathbf{0.920}$ $\mathsmaller{\pm 0.0009}$ \\
\midrule
\wire{} & $0.570$ $\mathsmaller{\pm 0.0017}$ & $0.695$ $\mathsmaller{\pm 0.0019}$ & $0.770$ $\mathsmaller{\pm 0.0040}$ \\
+best & $0.677$ $\mathsmaller{\pm 0.0008}$ & $0.787$ $\mathsmaller{\pm 0.0007}$ & $0.864$ $\mathsmaller{\pm 0.0014}$ \\
+\fresh{} & $0.000$ $\mathsmaller{\pm 0.0000}$ & $0.000$ $\mathsmaller{\pm 0.0000}$ & $0.000$ $\mathsmaller{\pm 0.0000}$ \\
+\our{} & $\mathbf{0.674}$ $\mathsmaller{\pm 0.0006}$ & $\mathbf{0.786}$ $\mathsmaller{\pm 0.0008}$ & $\mathbf{0.862}$ $\mathsmaller{\pm 0.0018}$ \\
\bottomrule
\end{tabular}

 \end{sc}
    \end{small}
\end{table}

\subsection{Cross-domain generalization of \our{}}
\label{sec:cross_domain}

To further validate \our{}, we perform a cross-domain validation in \cref{tab:sec_cross_domain} on the benchmark used for \fresh{} \cite{fresh}. We use \our{} with the same 10-image calibration dataset as used in \cref{sec:experiments}. We observe an increase in performance, showing strong generalization capabilities of \our{}. Particularly interesting is the particularly big performance gap on Kodak and Wiki Art datasets - those datasets are characterized by lower resolutions of images, which may indicate a connection between resolution and optimal \freqparam{} selection strategy.

\begin{table}
    \centering
    \caption{Evaluation of \our{} on benchmark used by \fresh{}. Results for baselines and \fresh{} are taken from the \fresh{} paper, while results for \our{} were computed by us over 3 seeds. Overall, we observe an increase in performance, showing strong generalization capabilities of \our{}.}
    \label{tab:sec_cross_domain}
    \vspace{5pt}
    \begin{tabular}{@{\hspace{2.5mm}}l@{\hspace{2.5mm}}c@{\hspace{2.5mm}}c@{\hspace{2.5mm}}c@{\hspace{2.5mm}}c@{\hspace{2.5mm}}c@{\hspace{2.5mm}}c@{\hspace{2.5mm}}}
    \toprule
 \small PSNR $\uparrow$    & \small Average & \small  \chest{} & \small  \ffhqcropped{} & \small  \ffhq{} & \small  \small  \kodak{} & \small \art{} \\
    \midrule
    \small \siren{} & \small $33.85$ $\mathsmaller{\pm 0.01}$ & \small $37.35$ $\mathsmaller{\pm {>} 0.01}$ & \small $37.54$ $\mathsmaller{\pm 0.04}$ & \small $34.32$ $\mathsmaller{\pm 0.01}$ & \small $31.60$ $\mathsmaller{\pm 0.03}$ & \small $28.45$ $\mathsmaller{\pm 0.01}$ \\
    \small +\fresh{} & \small $34.62$ $\mathsmaller{\pm 0.01}$ & \small $37.99$ $\mathsmaller{\pm 0.01}$ & \small $39.11$ $\mathsmaller{\pm 0.01}$ & \small $\mathbf{35.40}$ $\mathsmaller{\pm 0.01}$ & \small $31.78$ $\mathsmaller{\pm 0.02}$ & \small $28.80$ $\mathsmaller{\pm 0.01}$ \\
    \small +\our{} & \small $\mathbf{36.80}$ $\mathsmaller{\pm 0.01}$ & \small $\mathbf{38.20}$ $\mathsmaller{\pm 0.05}$ & \small $\mathbf{39.91}$ $\mathsmaller{\pm 0.02}$ & \small $35.15$ $\mathsmaller{\pm 0.02}$ & \small $\mathbf{36.99}$ $\mathsmaller{\pm 0.02}$ & \small $\mathbf{33.73}$ $\mathsmaller{\pm 0.02}$ \\

    \midrule
    \small \ffeatures{} & \small $32.12$ $\mathsmaller{\pm 0.01}$ & \small $36.96$ $\mathsmaller{\pm 0.03}$ & \small $35.01$ $\mathsmaller{\pm 0.04}$ & \small $32.65$ $\mathsmaller{\pm 0.01}$ & \small $28.84$ $\mathsmaller{\pm 0.04}$ & \small $27.15$ $\mathsmaller{\pm 0.01}$ \\
    \small +\fresh{} & \small $33.45$ $\mathsmaller{\pm 0.02}$ & \small $37.77$ $\mathsmaller{\pm 0.04}$ & \small $36.81$ $\mathsmaller{\pm 0.06}$ & \small $\mathbf{34.62}$ $\mathsmaller{\pm 0.01}$ & \small $30.06$ $\mathsmaller{\pm 0.01}$ & \small $28.01$ $\mathsmaller{\pm 0.02}$ \\
    \small +\our{} & \small $\mathbf{35.09}$ $\mathsmaller{\pm 0.11}$ & \small $\mathbf{38.20}$ $\mathsmaller{\pm 0.12}$ & \small $\mathbf{37.66}$ $\mathsmaller{\pm 0.11}$ & \small $34.40$ $\mathsmaller{\pm 0.04}$ & \small $\mathbf{33.51}$ $\mathsmaller{\pm 0.21}$ & \small $\mathbf{31.65}$ $\mathsmaller{\pm 0.16}$ \\

    \midrule
    \small \finer{} & \small $35.11$ $\mathsmaller{\pm {>} 0.01}$ & \small $38.63$ $\mathsmaller{\pm 0.02}$ & \small $40.45$ $\mathsmaller{\pm 0.01}$ & \small $\mathbf{36.48}$ $\mathsmaller{\pm 0.02}$ & \small $31.40$ $\mathsmaller{\pm 0.02}$ & \small $28.57$ $\mathsmaller{\pm 0.01}$ \\
    \small +\fresh{} & \small $35.03$ $\mathsmaller{\pm 0.01}$ & \small $38.51$ $\mathsmaller{\pm 0.04}$ & \small $40.31$ $\mathsmaller{\pm 0.07}$ & \small $\mathbf{36.48}$ $\mathsmaller{\pm 0.01}$ & \small $31.31$ $\mathsmaller{\pm 0.03}$ & \small $28.54$ $\mathsmaller{\pm 0.01}$ \\
    \small +\our{} & \small $\mathbf{37.63}$ $\mathsmaller{\pm 0.02}$ & \small $\mathbf{39.10}$ $\mathsmaller{\pm 0.07}$ & \small $\mathbf{41.54}$ $\mathsmaller{\pm 0.02}$ & \small $36.09$ $\mathsmaller{\pm 0.01}$ & \small $\mathbf{37.35}$ $\mathsmaller{\pm 0.01}$ & \small $\mathbf{34.04}$ $\mathsmaller{\pm 0.02}$ \\

    \midrule
    \small \wire{} & \small $\mathbf{33.54}$ $\mathsmaller{\pm 0.02}$ & \small $\mathbf{37.96}$ $\mathsmaller{\pm 0.02}$ & \small $\mathbf{38.13}$ $\mathsmaller{\pm 0.04}$ & \small $\mathbf{35.04}$ $\mathsmaller{\pm 0.01}$ & \small $28.95$ $\mathsmaller{\pm 0.06}$ & \small $27.62$ $\mathsmaller{\pm 0.01}$ \\
    \small +\our{} & \small $33.48$ $\mathsmaller{\pm 0.07}$ & \small $36.81$ $\mathsmaller{\pm 0.02}$ & \small $36.52$ $\mathsmaller{\pm 0.02}$ & \small $33.03$ $\mathsmaller{\pm 0.01}$ & \small $\mathbf{31.58}$ $\mathsmaller{\pm 0.01}$ & \small $\mathbf{29.48}$ $\mathsmaller{\pm 0.38}$ \\
\bottomrule
\end{tabular}
\end{table}

\subsection{Additional reconstruction metrics}
\label{sec:additional_metrics}

\Cref{tab:freq_param_prediction_ssim}
presents results using SSIM (Structural Similarity Index) metric which captures
perceptual quality more closely than PSNR.
\our{} exceeds or matches the performance of \fresh{} also when considering this metric.

\subsection{Network matching}
\label{sec:matching_extras}

\Cref{tab:network_matching_clipped} lists the specific frequency parameters selected by
our matching algorithm ($\operatorname{Fit}_{\text{ref}}(\cdot)$).
Note that we limited our matching to parameters around the default values used for
the respective models, which were included in our general grid search.
While this restricted our search space for matching, it is not an inherent limitation of the method,
and successful matching is achievable in all cases.

\begin{figure}[t!]
    \centering
    \includegraphics[width=95mm]{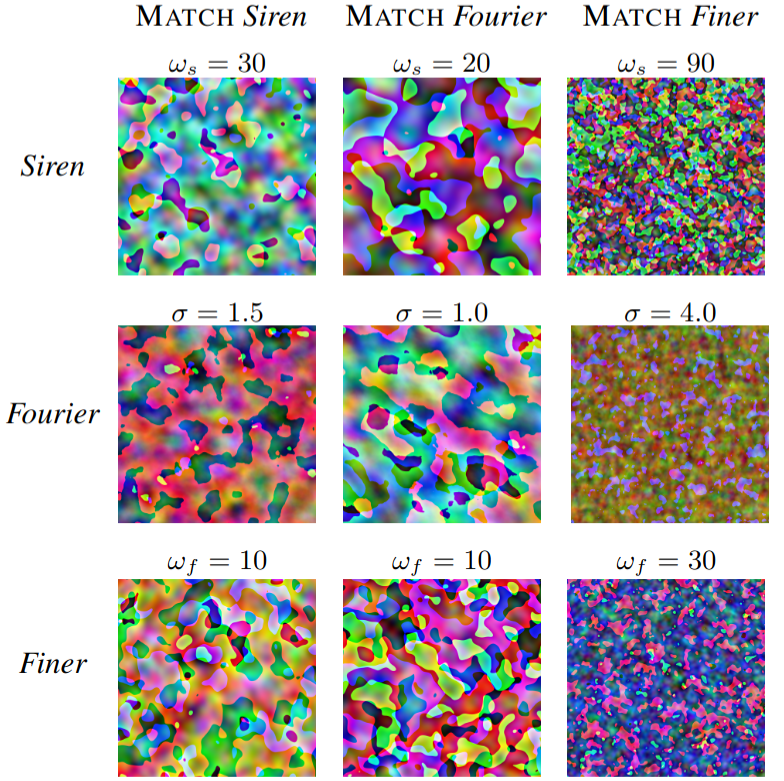}
    \caption{Network matching visualization: each row corresponds to a model
    architecture. 
    The columns display: (1) Matched to \siren{}, (2) Matched to \ffeatures{}, (3) Matched to \finer{}
    (diagonal represents reference configurations).
    The parameter search for this visualization was denser than for the main results.
    The value of \freqparam{} of the matched model is shown above each image.
    Observe that the outputs of unmatched models differ significantly but become
    similar after matching (images within a column are visually similar).
    }
    \label{fig:network_matching_viz}
\end{figure}

Additionally, we illustrate the outputs of matched models in \cref{fig:network_matching_viz}.

\begin{table}[t]
\caption{Parameters selected during frequency matching.
We report the selected parameter and the matching error
    (difference in SEC between reference and matched models).
High errors (e.g., matching to \wire{}, or matching \wire{} to other models) indicate that the target
    architecture could not replicate the reference's spectral bias within our search space.}
\label{tab:network_matching_clipped}
\vskip 0.15in
    \begin{center}
    \begin{small}
    \begin{sc}
    \begin{tabular}{lrrrrrr}
\toprule
& \multicolumn{2}{c}{S} & \multicolumn{2}{c}{M} & \multicolumn{2}{c}{L} \\
\cmidrule(lr){2-3} \cmidrule(lr){4-5} \cmidrule(lr){6-7}
 & Matched & SEC  & Matched & SEC  & Matched & SEC  \\
 & Parameter &  Error & Parameter &  Error & Parameter &  Error \\
\midrule
$\operatorname{Fit}_{\omega_f=30}(\omega_s)$ & 80.0 & 0.20 & 90.0 & 0.70 & 90.0 & 0.80 \\
$\operatorname{Fit}_{\omega_f=30}(\sigma)$ & 4 & 0.70 & 4 & 0.20 & 4 & 0.10 \\
$\operatorname{Fit}_{\omega_f=30}(\omega_w)$ & 10 & 7.40 & 10 & 45.10 & 10 & 190.50 \\
\midrule
$\operatorname{Fit}_{\omega_w=10}(\omega_f)$ & 50.0 &0.40 & 100 &14.10 & 100 &157.20 \\
$\operatorname{Fit}_{\omega_w=10}(\omega_s)$ & 110 &3.00 & 110 &41.20 & 110 &186.40 \\
$\operatorname{Fit}_{\omega_w=10}(\sigma)$ & 6 &0.40 & 9 &28.60 & 9 &172.40 \\
\midrule
$\operatorname{Fit}_{\omega_s=30}(\omega_f)$ & 10 & 0.50 & 10 & 0.30 & 10 & 0.00 \\
$\operatorname{Fit}_{\omega_s=30}(\sigma)$ & 1 & 1.00 & 1 & 1.40 & 1 & 1.20 \\
$\operatorname{Fit}_{\omega_s=30}(\omega_w)$ & 10 & 14.50 & 10 & 53.40 & 10 & 199.80 \\
\midrule
$\operatorname{Fit}_{\sigma=1.0}(\omega_f)$ & 10 & 0.50 & 10 & 1.100 & 10 & 1.20 \\
$\operatorname{Fit}_{\sigma=1.0}(\omega_s)$ & 30 & 1.00 & 30 & 1.40 & 30 & 1.20 \\
$\operatorname{Fit}_{\sigma=1.0}(\omega_w)$ & 10 & 15.50 & 10 & 54.80 & 10 & 201.00 \\
\bottomrule
\end{tabular}

    \end{sc}
    \end{small}
    \end{center}
\vskip -0.1in
\end{table}

\section{Ablation Studies}
\label{sec:ablation_studies}

To better understand which design choices are critical to performance,
we perform a series of ablation experiments.
We specifically focus on the formulation of the energy spectrum (see \cref{eq:spectrum_full}),
where in the main method we consider the DC component as always equal to 0 ($\power(A, 0)=0$),
following the example of \cite{fresh},
as the DC component can have a significant effect on the signal
while not making it harder nor easier to model.
When the DC component is not zeroed,
we label it as \our{}-DC.
We also consider using DFT coefficients directly,
as in \fresh{},
calculating the spectrum vector $\power$ (note: values of this vector can no longer be interpreted as the power of the original signal) as:

\begin{equation}
 \label{eq:spectrum_full_not_power}
 \power_\text{not power}(A, r) = \sum_{c=0}^{C-1} \sum_{\substack{(i,j) \in I \\
 \lfloor\sqrt{i^2+j^2}\rfloor=r}} |\mathcal{F}_{{\operatorname{Shift}(i,
 j)}}(A_c)|.
\end{equation}

Additionally, we consider a modification to SEC,
choosing a different statistic to convert $\power(A)$ into a scalar.
Specifically, we consider the median instead of the mean.
$\inherentFreqMath(A)$ is defined as the median of the energy spectrum.
Formally, it is the smallest frequency $f$ such that the cumulative energy up to $f$ accounts for at least 50\% of the total energy:
\begin{equation}
    \inherentFreqMath(A) = \min \left\{ f : \sum_{r=0}^{f} \normalPower(A, r) \ge 0.5 \right\}.
\end{equation}

\begin{figure}
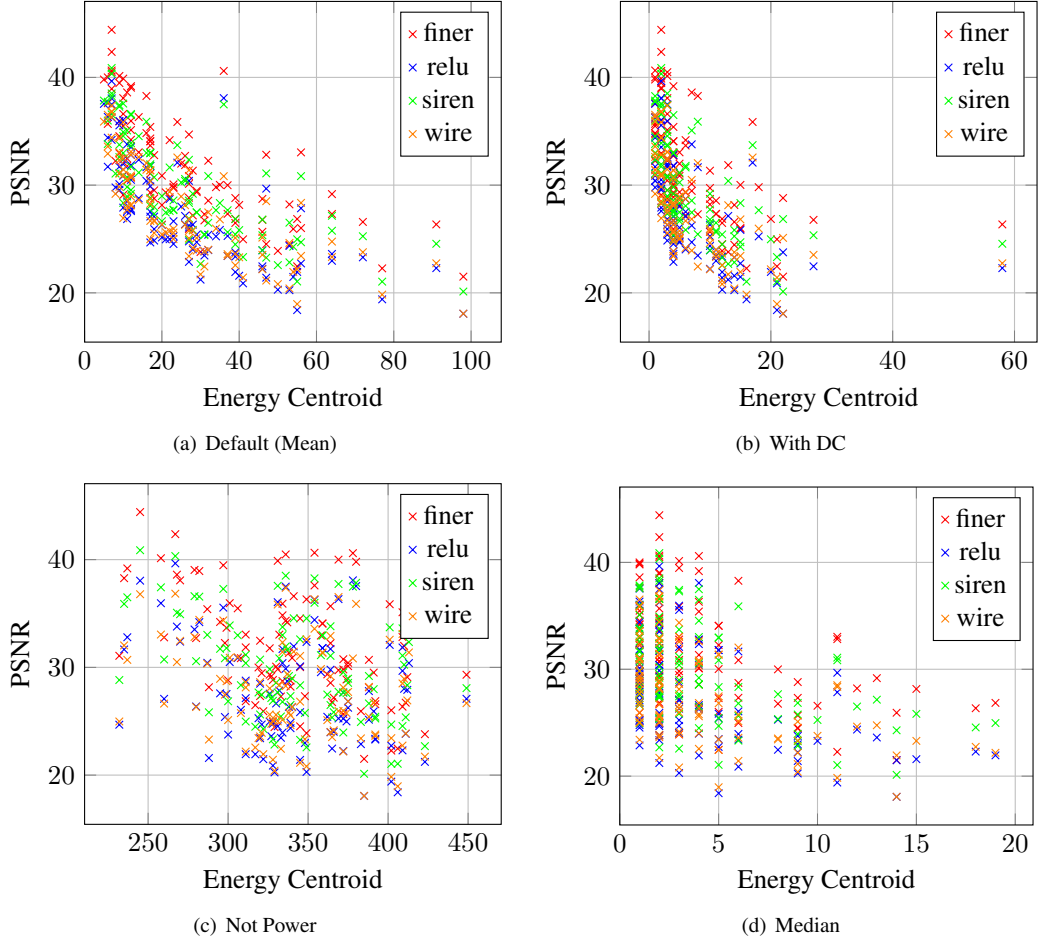

 \centering
 \subfigure[Default (Mean)]{
    \resizebox{0.48\textwidth}{!}{\input{experiment_results_mean/psnr_vs_freq_statistic}}
 }
 \hfill
 \subfigure[With DC]{
    \resizebox{0.48\textwidth}{!}{\input{experiment_results_mean_no_zero/psnr_vs_freq_statistic}}
 }
 \\
 \subfigure[Not Power]{
    \resizebox{0.48\textwidth}{!}{\input{experiment_results_mean_no_square/psnr_vs_freq_statistic}}
 }
 \hfill
 \subfigure[Median]{
    \resizebox{0.48\textwidth}{!}{\input{experiment_results_median/psnr_vs_freq_statistic}}
 }
 \caption{Relationship between \inherentfreq{} and reconstruction quality (PSNR) for Medium-sized models using default \freqparam{} values.
 We compare (a) our default configuration (Mean) against variations:
 (b) including the DC component,
 (c) using non-squared spectrum, and
 (d) using Median instead of Mean.}
 \label{fig:psnr_ablation_combined}
\end{figure}

We present results from this ablation study in
\cref{tab:psnr_correlation_ablation_combined}
and \cref{fig:psnr_ablation_combined}.
First, we observe that the prediction of optimal frequency parameters and
the resulting PSNR are not greatly affected by the specific choice of the
statistic function used to calculate the centroid (see \cref{tab:psnr_ablation}).
However, the correlation between \inherentfreq{} and the final reconstruction
quality (PSNR) degrades when the energy spectrum is not squared, when the
DC component is included, or when median is used instead of mean
(see \cref{tab:psnr_correlation_ablation_combined}).
In the case of including the DC component, this degradation likely occurs
because the high magnitude of the DC term compresses the range of
\inherentfreq{} values, pushing them closer to zero and causing significant
overlap between images with distinct frequency content,
as visualized in \cref{fig:psnr_ablation_combined} (b).

\begin{table}[ht]
 \centering
 \caption{Comparison of reconstruction quality (PSNR) for models configured via our method versus baseline defaults and \fresh{}.}
 \vskip 0.15in
 \begin{center}
    \begin{small}
 \begin{sc}
   \begin{tabular}{lccc}
\toprule
psnr $\uparrow$ & S & M & L \\
\midrule
\siren{} & $26.03$ $\mathsmaller{\pm 0.01}$ & $29.97$ $\mathsmaller{\pm 0.01}$ & $32.06$ $\mathsmaller{\pm 0.10}$ \\
+\our{} & $26.61$ $\mathsmaller{\pm 0.01}$ & $31.25$ $\mathsmaller{\pm 0.02}$ & $34.49$ $\mathsmaller{\pm 0.05}$ \\
+\our{}-median & $26.61$ $\mathsmaller{\pm 0.01}$ & $31.25$ $\mathsmaller{\pm 0.02}$ & $34.44$ $\mathsmaller{\pm 0.08}$ \\
+\our{}-not-power & $26.60$ $\mathsmaller{\pm 0.01}$ & $31.23$ $\mathsmaller{\pm 0.01}$ & $34.54$ $\mathsmaller{\pm 0.02}$ \\
+\our{}-DC & $26.62$ $\mathsmaller{\pm 0.01}$ & $31.25$ $\mathsmaller{\pm 0.02}$ & $34.52$ $\mathsmaller{\pm 0.04}$ \\
\midrule
\ffeatures{} & $24.72$ $\mathsmaller{\pm 0.04}$ & $27.68$ $\mathsmaller{\pm 0.11}$ & $29.01$ $\mathsmaller{\pm 0.03}$ \\
+\our{} & $26.35$ $\mathsmaller{\pm 0.07}$ & $30.45$ $\mathsmaller{\pm 0.14}$ & $31.74$ $\mathsmaller{\pm 0.08}$ \\
+\our{}-median & $26.21$ $\mathsmaller{\pm 0.01}$ & $30.52$ $\mathsmaller{\pm 0.15}$ & $31.65$ $\mathsmaller{\pm 0.01}$ \\
+\our{}-not-power  & $26.32$ $\mathsmaller{\pm 0.01}$ & $30.54$ $\mathsmaller{\pm 0.03}$ & $31.93$ $\mathsmaller{\pm 0.13}$ \\
+\our{}-DC & $26.31$ $\mathsmaller{\pm 0.02}$ & $30.48$ $\mathsmaller{\pm 0.10}$ & $31.75$ $\mathsmaller{\pm 0.07}$ \\
\midrule
\finer{} & $27.26$ $\mathsmaller{\pm 0.02}$ & $31.86$ $\mathsmaller{\pm 0.01}$ & $34.23$ $\mathsmaller{\pm 0.01}$ \\
+\our{} & $27.21$ $\mathsmaller{\pm 0.02}$ & $31.99$ $\mathsmaller{\pm 0.01}$ & $35.81$ $\mathsmaller{\pm 0.07}$ \\
+\our{}-median & $27.23$ $\mathsmaller{\pm 0.02}$ & $31.99$ $\mathsmaller{\pm 0.01}$ & $35.76$ $\mathsmaller{\pm 0.05}$ \\
+\our{}-not-power  & $27.24$ $\mathsmaller{\pm 0.02}$ & $32.00$ $\mathsmaller{\pm 0.01}$ & $35.80$ $\mathsmaller{\pm 0.04}$ \\
+\our{}-DC & $27.23$ $\mathsmaller{\pm 0.03}$ & $31.99$ $\mathsmaller{\pm 0.00}$ & $35.75$ $\mathsmaller{\pm 0.03}$ \\
\midrule
\wire{} & $24.39$ $\mathsmaller{\pm 0.02}$ & $27.70$ $\mathsmaller{\pm 0.04}$ & $28.67$ $\mathsmaller{\pm 0.08}$ \\
+\our{} & $26.07$ $\mathsmaller{\pm 0.01}$ & $29.43$ $\mathsmaller{\pm 0.03}$ & $31.58$ $\mathsmaller{\pm 0.17}$ \\
+\our{}-median & $26.06$ $\mathsmaller{\pm 0.02}$ & $29.44$ $\mathsmaller{\pm 0.03}$ & $31.64$ $\mathsmaller{\pm 0.11}$ \\
+\our{}-not-power  & $26.07$ $\mathsmaller{\pm 0.02}$ & $29.42$ $\mathsmaller{\pm 0.02}$ & $31.61$ $\mathsmaller{\pm 0.07}$ \\
+\our{}-DC & $26.05$ $\mathsmaller{\pm 0.01}$ & $29.44$ $\mathsmaller{\pm 0.03}$ & $31.62$ $\mathsmaller{\pm 0.13}$ \\
\bottomrule
\end{tabular}

 \end{sc}
    \end{small}
 \end{center}
 \label{tab:psnr_ablation}
\end{table}

    \begin{table}[t]
 \centering
 \caption{Pearson correlation between \inherentfreq{} and reconstruction quality (PSNR).
 We compare (a) our default configuration (Mean) against variations:
 (b) including the DC component,
 (c) using non-squared spectrum, and
 (d) using Median instead of Mean.}
 \label{tab:psnr_correlation_ablation_combined}
 \vskip 0.15in
 \hfill
 \subfigure[Default (Mean)]{
    {\begin{tabular}{lccc}
\toprule
Corr $\downarrow$ & S & M & L \\
\midrule
siren & -0.866 & -0.822 & -0.760 \\
relu & -0.839 & -0.834 & -0.845 \\
finer & -0.861 & -0.796 & -0.739 \\
wire & -0.849 & -0.843 & -0.695 \\
\bottomrule
\end{tabular}

}
 }
 \hfill
 \subfigure[With DC]{
    {\begin{tabular}{llll}
\toprule
 & S & M & L \\
\midrule
siren & -0.792 & -0.760 & -0.690 \\
relu & -0.797 & -0.771 & -0.784 \\
finer & -0.782 & -0.741 & -0.674 \\
wire & -0.784 & -0.764 & -0.681 \\
\bottomrule
\end{tabular}

}
 }
 \hfill
 \\
 \hfill
 \subfigure[Not Power]{
    {\begin{tabular}{llll}
\toprule
 & S & M & L \\
\midrule
siren & -0.132 & -0.325 & -0.432 \\
relu & -0.033 & -0.214 & -0.301 \\
finer & -0.166 & -0.366 & -0.463 \\
wire & -0.031 & -0.200 & -0.292 \\
\bottomrule
\end{tabular}

}
 }
 \hfill
 \subfigure[Median]{
    {\begin{tabular}{llll}
\toprule
 & S & M & L \\
\midrule
siren & -0.547 & -0.467 & -0.404 \\
relu & -0.574 & -0.501 & -0.503 \\
finer & -0.538 & -0.440 & -0.389 \\
wire & -0.578 & -0.519 & -0.420 \\
\bottomrule
\end{tabular}

}
 }
 \hfill
 \vskip -0.1in
    \end{table}

  \begin{table}[t]
 \caption{Accuracy of \fresh{} and \our{} in predicting the optimal frequency parameter.}
 \label{tab:psnr_accuracy_ablation}
 \vskip 0.15in
 \begin{center}
    \begin{small}
 \begin{sc}
   \begin{tabular}{lccc}
\toprule
psnr $\uparrow$ & L & M & S \\
\midrule
\siren{} + \our{} & $0.64$  & $0.21$  & $0.36$  \\
+ \our{}-median & $0.50$ & $0.26$ & $0.43$ \\
+\our{}-not-power & $0.84$  & $0.29$  & $0.46$  \\
+\our{}-DC & $0.77$  & $0.23$  & $0.41$  \\
\midrule
\ffeatures{} + \our{} & $0.38$  & $0.27$  & $0.17$  \\
+\our{}-not-power & $0.54$  & $0.33$  & $0.18$  \\
+\our{}-DC & $0.27$  & $0.19$  & $0.16$  \\
\midrule
\finer{} + \our{} & $0.44$  & $0.19$  & $0.54$  \\
+\our{}-not-power & $0.56$  & $0.13$  & $0.47$  \\
+\our{}-DC & $0.44$  & $0.20$  & $0.54$  \\
\midrule
\wire{} + \our{} & $0.44$  & $0.52$  & $0.52$  \\
+\our{}-not-power & $0.62$  & $0.57$  & $0.43$  \\
+\our{}-DC & $0.54$  & $0.54$  & $0.54$  \\
\bottomrule
\end{tabular}

 \end{sc}
    \end{small}
 \end{center}
 \vskip -0.1in
    \end{table}

\section{\finer{} Initialization}
\label{sec:finer_init_issue}

In this section, we investigate the effect of model width on the spectral properties of \finer{}
to explain the behavior noted in \cref{sec:effects_of_architecture_on_sec}.

We start by noting that the \siren{} architecture \cite{SIREN} employs a carefully designed initialization scheme
to control variance and limit the frequency increase within the network.
Specifically, the pre-activation values (inputs to the nonlinearity)
are expected to follow a standard normal distribution with unit variance.
Although \finer{} \cite{liu2024finer} adopts the same initialization strategy,
it breaks the assumptions made by \siren{} by introducing an input-dependent scaling term $|\mathbf{W} \mathbf{x} + \mathbf{b}| + 1$:

\begin{equation}
    \gamma_F(\mathbf{x}) = \sin\Bigl(\omega_f (|\mathbf{W} \mathbf{x} +
    \mathbf{b}| + 1)(\mathbf{W} \mathbf{x} +
    \mathbf{b})\Bigr).
\end{equation}

\begin{figure}[ht]
    \centering
    \subfigure[Width 128]{
 \includegraphics[width=0.48\textwidth]{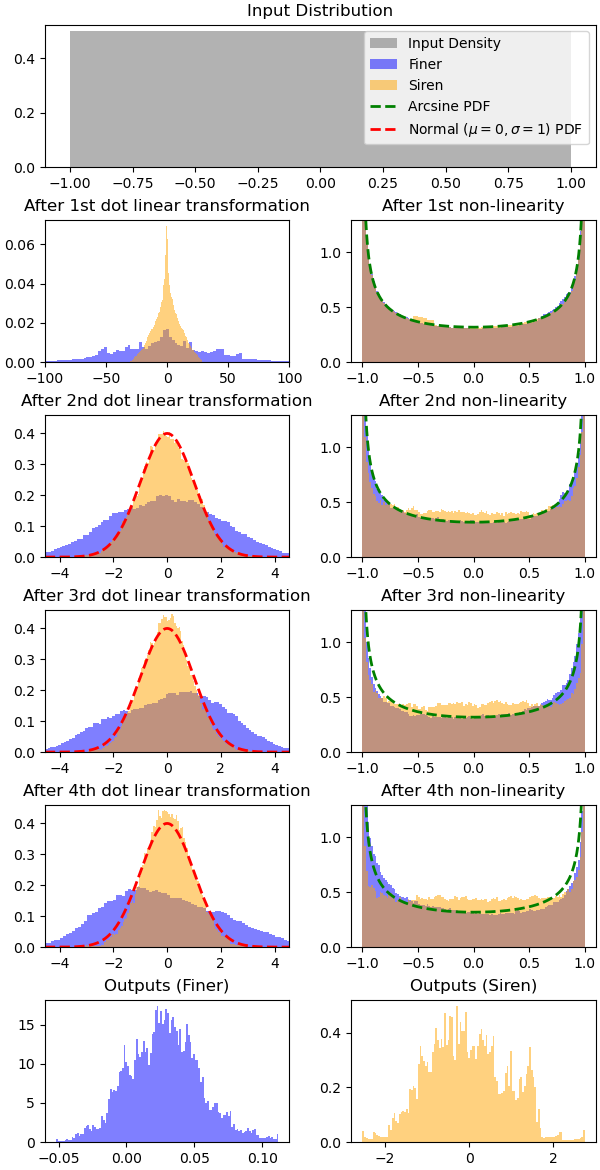}
    }
    \hfill
    \subfigure[Width 512]{
 \includegraphics[width=0.48\textwidth]{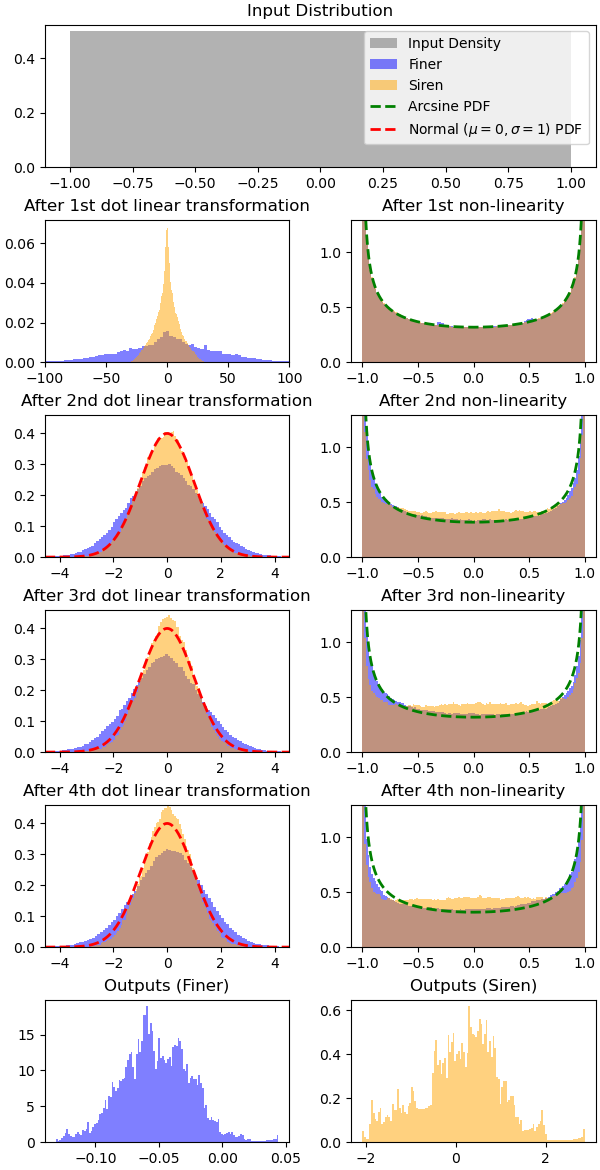}
    }
    \caption{Distributions of pre- and post-activation (sine) values within \finer{}
    for different widths.
    Narrower models (left) exhibit higher variance in pre-activation values
    compared to wider models (right).
    This increased variance in the inputs to the sine activation directly corresponds to higher frequencies,
    resulting in higher \inherentfreq{} for narrower and deeper architectures.}
    \label{fig:siren_finer_distribution}
\end{figure}

We show the effect of this term on the intermediate outputs of \finer{} in \cref{fig:siren_finer_distribution},
where we plot the distributions of activations at various stages in \finer{} and \siren{} networks.
We observe that the pre-sine activations of \finer{} exhibit higher variance than those of \siren{},
and this variance is width-dependent.
Specifically, the variance of activations increases as the model width decreases.
This increased variance in the inputs to the sine activation directly corresponds to higher frequencies,
explaining the increase in SEC for smaller models observed in \cref{fig:freq_vs_width_and_layers}.
We believe this uncovers an interesting research direction for the \finer{} architecture,
which would likely benefit from an initialization scheme tailored to its variable-periodic activations
that would avoid the unexpected influence of width on spectral bias.

\end{document}